\PassOptionsToPackage{table,dvipsnames}{xcolor}
\documentclass[runningheads]{llncs}

\usepackage{eccv}

\usepackage{eccvabbrv}

\usepackage{graphicx}
\usepackage{booktabs}

\usepackage[accsupp]{axessibility}  % Improves PDF readability for those with disabilities.

\usepackage{hyperref}

\usepackage{orcidlink}

\usepackage{placeins}   % \FloatBarrier to keep floats near their references
\usepackage{float}      % Provides [H] float placement
\usepackage{booktabs}
\usepackage{array}
\usepackage{multirow}
\usepackage[table]{xcolor}
\usepackage{pifont}

\usepackage{acro}
\usepackage{enumitem}
\usepackage{tikz}
\usetikzlibrary{backgrounds}
\usetikzlibrary{positioning}
\usetikzlibrary{calc,fit,arrows.meta,shapes.geometric}

\newcommand{\cmark}{\ding{51}} % check mark
\newcommand{\xmark}{\ding{55}} % cross mark

\DeclareAcronym{GNSS}{short = GNSS, long = Global Navigation Satellite System}
\DeclareAcronym{UAV}{short = UAV, long = Unmanned Aerial Vehicle}
\DeclareAcronym{DOP}{short = DOP, long = Digital Orthophoto}
\DeclareAcronym{DSM}{short = DSM, long = Digital Surface Model}
\DeclareAcronym{SLAM}{short = SLAM, long = Simultaneous Localization and Mapping}
\DeclareAcronym{PnP}{short = PnP, long = Perspective-n-Point}
\DeclareAcronym{LoD}{short = LoD, long = Level of Detail}
\DeclareAcronym{LiDAR}{short = LiDAR, long = Light Detection and Ranging}
\DeclareAcronym{MAP}{short = MAP, long = Maximum A Posteriori}
\DeclareAcronym{IMU}{short = IMU, long = Inertial Measurement Unit}
\DeclareAcronym{VO}{short = VO, long = Visual Odometry}
\DeclareAcronym{MAV}{short = MAV, long = Micro Aerial Vehicles}
\DeclareAcronym{DoF}{short = DoF, long = degrees of freedom}
\DeclareAcronym{FOV}{short = FOV, long = field of view}
\DeclareAcronym{AGL}{short = AGL, long = Above Ground Level}
\DeclareAcronym{SfM}{short = SfM, long = Structure from Motion}

\begin{document}

\title{OrthoTrack: Continuous 6-DoF UAV Trajectory Estimation Anchored in Public Orthophotos} 

\titlerunning{OrthoTrack}

\author{Oussema Dhaouadi\inst{1,2,3,4}\thanks{Corresponding author: oussema.dhaouadi@tum.de}\orcidlink{0009-0008-6842-5220} \and
Zuria Bauer\inst{1}\orcidlink{0000-0001-8447-2344} \and
Johannes Meier\inst{2,4}\orcidlink{0009-0000-2227-8271} \\
Olaf Wysocki\inst{3}\orcidlink{0000-0002-0016-0229} \and
Marc Pollefeys\inst{1,5}\orcidlink{0000-0003-2448-2318} \and
Daniel Cremers\inst{2,4}\orcidlink{0000-0002-3079-7984}}

\authorrunning{O.~Dhaouadi et al.}

\institute{
\textsuperscript{1}\,ETH Zurich\;\;\;\;
\textsuperscript{2}\,TU Munich\;\;\;\;
\textsuperscript{3}\,University of Cambridge\;\;\;\;
\textsuperscript{4}\,MCML\;\;\;\;
\textsuperscript{5}\,Microsoft \\
\url{https://orthotrack.ethz.ch}
}

\maketitle

\begin{abstract}\sloppy
Continuous 6-DoF pose estimation is essential for autonomous UAV operations. Yet, existing visual odometry and SLAM methods accumulate drift and yield only relative, up-to-scale trajectories. Single-frame geo-localization, in turn, discards temporal continuity and remains too slow for real-time use.
We present OrthoTrack, a training-free system that estimates continuous 6-DoF UAV trajectories using only publicly available orthophotos and surface models as a map prior. OrthoTrack matches keyframes against the orthophoto and lifts correspondences to metric 3D via the surface model. It then propagates these map-anchored correspondences to intermediate frames with optical flow, producing absolute, metrically scaled poses at every frame without GPS or post-hoc alignment.
We also introduce the MovingDrone Dataset, a large-scale benchmark pairing photorealistic UAV sequences with dense 6-DoF ground truth and co-registered multi-modal geodata including multi-temporal orthophotos. On MovingDrone and real-world benchmarks, OrthoTrack runs in real time on a single GPU. It outperforms all baselines by a large margin, even those receiving oracle scale and alignment. By relying on publicly available geodata, OrthoTrack enables deployment to new regions without site-specific adaptation.
\end{abstract}

\begin{figure*}[ht]
    \centering
    \includegraphics[width=\linewidth]{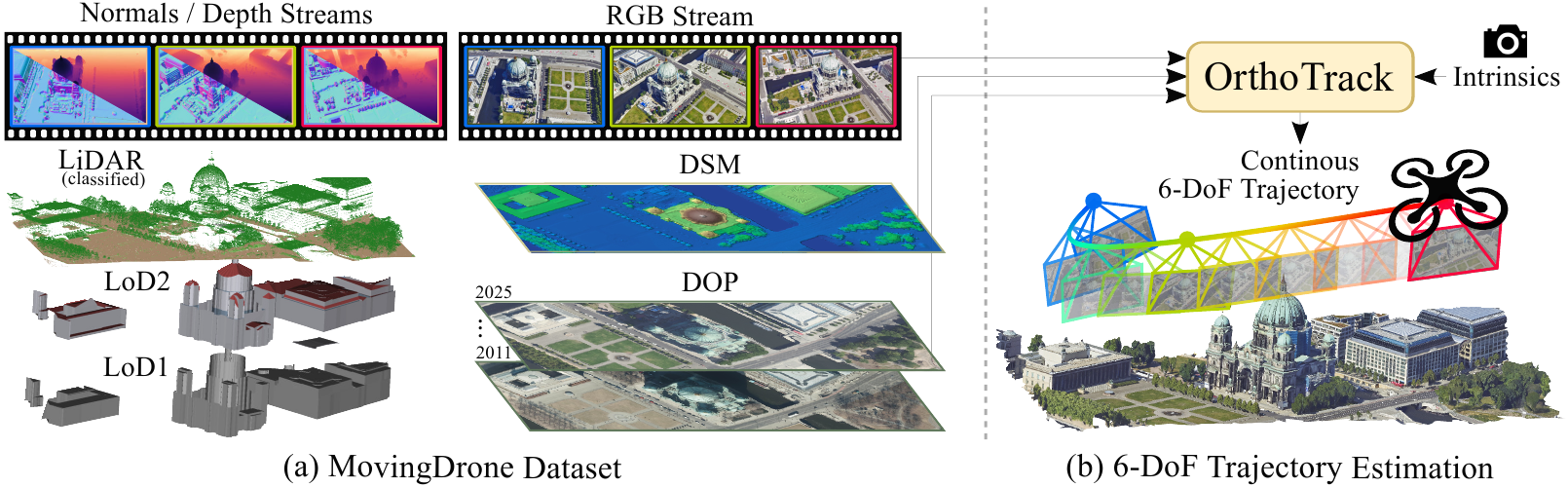}
    \caption{\textbf{OrthoTrack Estimates Continuous 6-DoF UAV Trajectories From Public Geodata.} (a)~Our MovingDrone Dataset pairs photorealistic UAV sequences with dense 6-DoF ground truth and multi-modal geodata supporting diverse aerial perception tasks (LiDAR, LoD1/2 meshes, DSM, multi-year DOPs, depth, and normals). (b)~OrthoTrack matches UAV frames against the DOP, lifts correspondences to 3D via the DSM, and outputs metrically scaled poses at every frame.}
    \label{fig:teaser}
\vspace{-0.5\baselineskip}
\end{figure*}
\section{Introduction}
\label{sec:intro}

Autonomous \acp{UAV} are increasingly deployed for infrastructure inspection~\cite{inspection_18}, emergency response~\cite{rescue_15}, and environmental monitoring~\cite{wildlife_monitoring_16}, all requiring accurate, continuous 6-\ac{DoF} pose estimation. While \ac{GNSS} provides coarse position, it offers no orientation and degrades or fails in urban canyons and signal-denied environments. Vision-based localization is a natural alternative, yet existing paradigms have fundamental limitations. For instance, \ac{VO} and \ac{SLAM}\cite{orbslam3, teed2021droid, teed2024deep} track features but accumulate unbounded drift without global anchors. Image retrieval\cite{anyvisloc_25, university_1652_20, sun2025cross} provides limited metric accuracy and requires large databases. Mesh-based methods~\cite{meshloc_22, render_and_compare_23} achieve high accuracy but depend on dense 3D models that are costly to maintain and prohibitive for continuous video. Furthermore, \ac{LoD} city-model approaches~\cite{zhu2024lodloc,zhu2025lodloc2} are compact and privacy-preserving yet restricted to structured urban geometry.

A practical alternative is to localize against publicly available \acp{DOP} and \acp{DSM}, which are metrically accurate, globally registered geodata that government agencies publish at centimeter resolution~\cite{dhaouadi2025ortholoc,anyvisloc_25}. Recent dense matchers~\cite{edstedt2025roma2, wang2024efficient} have made cross-view matching between oblique \ac{UAV} imagery and top-down orthophotos feasible, achieving sub-meter accuracy on individual frames~\cite{dhaouadi2025ortholoc,anyvisloc_25}. However, extending this to continuous video exposes two shortcomings: each dense match is too computationally expensive for frame-rate operation, and independent frames can produce isolated outliers with no mechanism for temporal recovery.

To address both shortcomings, we present OrthoTrack, a training-free system for continuous 6-\ac{DoF} \ac{UAV} trajectory estimation from publicly available \ac{DOP} and \ac{DSM} data (\cref{fig:teaser}). The key insight is that geodata-based localization and optical flow tracking are complementary: the former provides drift-free absolute anchoring, the latter temporal continuity at frame-rate speed. OrthoTrack geo-localizes sparse keyframes against the \ac{DOP} and lifts correspondences to metric 3D positions via the \ac{DSM}. It then propagates these anchored 2D-3D correspondences to intermediate frames using optical flow. Every pose is thereby globally registered and metrically scaled by construction, without the post-hoc alignment that \ac{VO}/\ac{SLAM} methods require.

We also introduce the MovingDrone Dataset, a large-scale benchmark for continuous \ac{UAV} localization with dense 6-\ac{DoF} ground truth (\cref{fig:teaser}). To our knowledge, it is the first to combine continuous video with per-frame metric poses and multi-temporal geodata. Prior datasets offer only single-frame evaluation~\cite{dhaouadi2025ortholoc} or at most two map years~\cite{anyvisloc_25}. We render photorealistic video from a government-grade textured mesh, yielding pixel-perfect poses with a genuine cross-domain gap since the mesh and reference geodata stem from independent surveys. It covers 21~Berlin regions with \ac{DOP} imagery from up to 14 years, and the pipeline extends to any city with publicly available geodata.

Our contributions are threefold. (1)~OrthoTrack, a training-free, matcher- and flow-agnostic system for continuous 6-\ac{DoF} \ac{UAV} trajectory estimation that produces metrically scaled absolute poses without GPS, task-specific training, or post-hoc alignment. (2)~The MovingDrone Dataset, a large-scale benchmark with dense ground truth, co-registered multi-modal geodata, and a scalable generation pipeline. (3)~State-of-the-art results against \ac{VO}/\ac{SLAM}, 3D foundation models, and \ac{UAV} localization baselines at real-time throughput.
\section{Related Work}
\label{sec:related_work}

\noindent\textbf{UAV Sparse Localization.}
Single-image UAV localization estimates the 6-\ac{DoF} camera pose of an individual aerial frame against a pre-existing map. Structure-based approaches match against \ac{SfM} point clouds or textured meshes~\cite{meshloc_22, render_and_compare_23}. UAVD4L~\cite{wu2023uavd4l} renders synthetic views from a textured city model for coarse retrieval and 6-\ac{DoF} estimation, but such reconstructions are expensive. \ac{LoD} city-model approaches exploit structured building geometries: LoD-Loc~\cite{zhu2024lodloc} aligns neural wireframes against \ac{LoD}3 models, while LoD-Loc~v2~\cite{zhu2025lodloc2} uses silhouette alignment on more widely available \ac{LoD}1/2 data. OrthoLoC~\cite{dhaouadi2025ortholoc} localizes against \acp{DOP} and \acp{DSM}, demonstrating that lightweight public geodata suffices for accurate 6-\ac{DoF} estimation. All these methods process frames independently and do not exploit temporal continuity. The cross-view matching they rely on has been advanced by dense matchers~\cite{edstedt2023dkm, edstedt2024roma, edstedt2025roma2}, especially with cross-domain training~\cite{wang2024efficient} and aerial-ground data~\cite{vuong2025aerialmegadepth}. OrthoTrack adopts RoMa v2~\cite{edstedt2025roma2} for this step.

\noindent\textbf{Visual Odometry and SLAM.}
Monocular \ac{VO} and \ac{SLAM} recover camera motion from frame-to-frame correspondences but produce trajectories with unknown metric scale. While loop closure and relocalization, as in ORB-SLAM3~\cite{orbslam3}, can reduce accumulated drift, they require revisiting previously observed geometry, which is rare in typical \ac{UAV} forward-flight scenarios. Recent learning-based methods address some of these shortcomings: DROID-SLAM~\cite{teed2021droid} performs differentiable dense bundle adjustment and DPVO~\cite{teed2024deep} offers a competitive speed-accuracy trade-off via recurrent patch updates, yet both still suffer from unbounded drift without loop closures and lack absolute metric scale. OrthoTrack addresses both by periodically re-anchoring to \ac{DSM}-derived georeferenced 3D positions, producing globally consistent, metrically scaled trajectories.

\noindent\textbf{Feed-Forward 3D Foundation Models.}
Recent feed-forward models predict 3D geometry directly from images, bypassing classical multi-view pipelines. While designed for general-purpose 3D understanding, they can in principle estimate camera trajectories from video, making them relevant baselines for \ac{UAV} pose estimation. DUSt3R~\cite{wang2024dust3r} pioneered pairwise point-map prediction from uncalibrated image pairs; follow-ups extend the paradigm to dense matching (MASt3R~\cite{leroy2024grounding}), feed-forward multi-view inference (MUSt3R~\cite{cabon2025must3r}), joint camera-depth-track prediction (VGGT~\cite{wang2025vggt}), and metric-scale reconstruction (Pow3R \cite{jang2025pow3r}, DA3~\cite{lin2025depth}, MapAnything~\cite{keetha2025mapanything}, Pi3~\cite{wang2025pi3}). Despite their generality, all these models produce local-frame reconstructions that require a similarity transform for georeferenced alignment, limiting their applicability to absolute \ac{UAV} pose estimation. In contrast, OrthoTrack directly anchors every prediction to the geodata reference frame, avoiding post-hoc alignment entirely. %We benchmark representative variants on aerial localization and depth estimation using our dataset.

% --- Comparison Table ---------------------------------------------------------
\begin{table*}[t]
    \centering
    \caption{\textbf{Comparison With Existing UAV Localization Benchmarks.}}
    \label{tab:dataset_comparison}
    \setlength{\tabcolsep}{1.5pt}
    \small
    \resizebox{\textwidth}{!}{%
    \begin{tabular}{@{}llcccccccccccccc@{}}
        \toprule
        Dataset & Year & Data & Scene & \#Loc & Frames & View & Alt. & fps & Depth & 6-DoF & DOP & DSM & LiDAR & Mesh & LoD \\
        \midrule
        University-1652~\cite{university_1652_20} & 2020 & Re+Sy\textsuperscript{R} & C & 72 & 50k & O & B & -- & \xmark & \xmark & 1$^\dagger$ & \xmark & \xmark & \xmark & \xmark \\
        VPAIR~\cite{vpair_22} & 2022 & Re & M & 1 & 2.7k & N & H & 1 & \cmark & \cmark & 1 & \xmark & \cmark & \xmark & \xmark \\
        ALTO~\cite{alto_22} & 2022 & Re & M & 1 & 15.4k & N & H & 20$^\S$ & \xmark & \cmark & 1 & \xmark & \cmark & \xmark & \xmark \\
        CrossLoc~\cite{crossloc_22} & 2022 & Re+Sy & U/N & 2 & 24k & B & L & -- & \cmark & \cmark & \xmark & \xmark & \xmark & \xmark & \cmark$^\ddagger$ \\
        UAVD4L~\cite{wu2023uavd4l} & 2024 & Re+Sy & U & 2 & 19k & B & L & 0.5 & \cmark & \cmark & \xmark & \cmark & \xmark & \cmark & \cmark$^\ddagger$ \\
        UAV-VisLoc~\cite{uav_visloc_24} & 2024 & Re & M & 11 & 6.7k & N & H & 0.2 & \xmark & \xmark & 1$^\dagger$ & \xmark & \xmark & \xmark & \xmark \\
        GTA-UAV~\cite{gta_uav_25} & 2025 & Sy\textsuperscript{G} & M & 1 & 33k & N & B & -- & \xmark & \cmark & \xmark & \xmark & \xmark & \xmark & \xmark \\
        OrthoLoC~\cite{dhaouadi2025ortholoc} & 2025 & Re & M & 47 & 16.4k & B & B & -- & \cmark & \cmark & 2 & \cmark & \xmark & \cmark & \xmark \\
        AnyVisLoc~\cite{anyvisloc_25} & 2025 & Re & M & 25 & 18k & B & B & -- & \xmark & \xmark & 1+1$^\dagger$ & \cmark & \xmark & \xmark & \xmark \\
        UAVScenes~\cite{wang2025uavscenes} & 2025 & Re & M & 4 & 120k & B & L & 10 & \xmark & \cmark & \xmark & \xmark & \cmark & \cmark & \xmark \\
        \midrule
        \textbf{MovingDrone (Ours)} & 2026 & Sy\textsuperscript{R} & M & 21 & 343k & B & B & 30 & \cmark & \cmark & $\leq$14 & \cmark & \cmark & \cmark & \cmark \\
        \bottomrule
    \end{tabular}}\par
    \vspace{0.25\baselineskip}
    {\raggedright\scriptsize Re=real, Sy\textsuperscript{R}=synthetic (photorealistic), Sy\textsuperscript{G}=synthetic (game engine); U=urban, C=campus, M=mixed, N=nature; N=nadir, O=oblique, B=both; L=low ($\leq$150\,m), H=high ($>$150\,m).\, $^\dagger$Satellite tiles, not surveyed orthophotos.\, $^\ddagger$Datasets extended to UAVD4L-LoD and Swiss-EPFL with LoD models in~\cite{zhu2024lodloc,zhu2025lodloc2}.\, $^\S$Full dataset never publicly released; only sparse competition subsets available.\par}
%\vspace{0\baselineskip}
\end{table*}
\noindent\textbf{UAV Localization Datasets.}
Existing benchmarks cover only a subset of the requirements for evaluating continuous map-based localization (\cref{tab:dataset_comparison}).
Several benchmarks~\cite{university_1652_20, gta_uav_25, vpair_22, alto_22, uav_visloc_24} target geo-localization or place recognition but provide either no metric 6-\ac{DoF} poses~\cite{university_1652_20,uav_visloc_24} or only sparse imagery~\cite{university_1652_20,crossloc_22,gta_uav_25,dhaouadi2025ortholoc,anyvisloc_25} unsuitable for sequential tracking.
Among datasets with 6-\ac{DoF} annotations, CrossLoc~\cite{crossloc_22} ships no geodata, UAVD4L~\cite{wu2023uavd4l} pairs mesh-derived geodata with only 0.5\,fps imagery, and AnyVisLoc~\cite{anyvisloc_25} provides single images with two map resolutions.
Closest to our setting, OrthoLoC~\cite{dhaouadi2025ortholoc} pairs \acp{DOP}/\acp{DSM} with 6-\ac{DoF} poses across 47 locations but supplies only sparse, independent frames, while UAVScenes~\cite{wang2025uavscenes} offers dense video with \ac{LiDAR} but ships neither \acp{DOP} nor \acp{DSM}.
To our knowledge, MovingDrone is the first dataset to combine dense video with a jointly aligned, multi-temporal suite of geodata modalities.
\section{Method}
\label{sec:method}
\begin{figure*}[ht]
    \centering
    \includegraphics[width=0.92\linewidth]{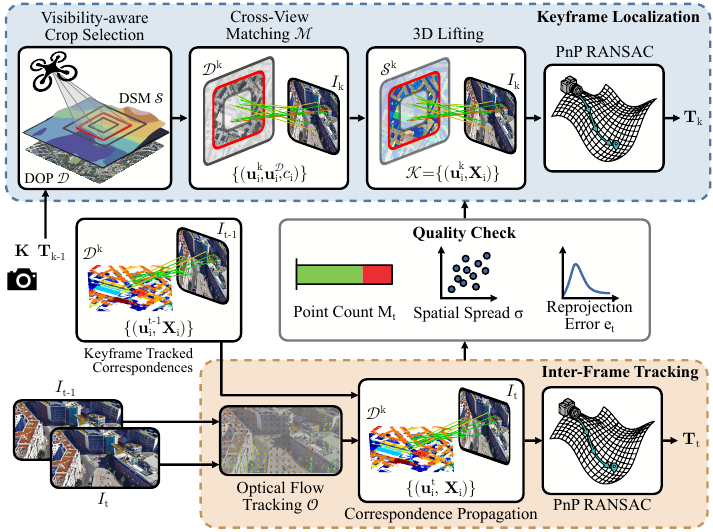}
    \caption{\textbf{OrthoTrack Pipeline.} Keyframe localization (top): a visibility-aware crop selects \ac{DOP} $\mathcal{D}^k$ and \ac{DSM} $\mathcal{S}^k$ from the previous pose. The dense matcher $\mathcal{M}$ matches keyframe $\mathbf{I}_k$ against $\mathcal{D}^k$, correspondences are lifted to 3D via $\mathcal{S}^k$, and PnP-RANSAC recovers the keyframe pose $\mathbf{T}_k$. Inter-frame tracking (bottom): optical flow $\mathcal{O}$ propagates keyframe-anchored correspondences from previous frame $\mathbf{I}_{t-1}$ to current frame $\mathbf{I}_t$ for a lightweight PnP solve. The adaptive quality monitor (center) triggers re-localization when point count, spatial spread, or reprojection error degrades.}
    \label{fig:pipeline}
%\vspace{-0.5\baselineskip}
\end{figure*}Given a \ac{UAV} video and a georeferenced map consisting of a \ac{DOP} and a \ac{DSM}, OrthoTrack estimates a 6-\ac{DoF} pose for every frame (\cref{fig:pipeline}). The system first localizes itself on the map without any external sensor. It then enters a loop that alternates between two modes: at keyframes, a dense cross-view match against the \ac{DOP} produces a globally anchored pose. Between keyframes, optical flow propagates the existing correspondences for lightweight pose updates. An adaptive quality monitor triggers new keyframes when tracking degrades. We begin with notation and then detail each component.

\subsection{Preliminaries}
\label{sec:preliminaries}

\noindent\textbf{Notation.}
We denote the \ac{UAV} video as $\{\mathbf{I}_t\}_{t=1}^{T}$ with known pinhole intrinsics $\mathbf{K} \in \mathbb{R}^{3 \times 3}$. All poses are world-to-camera: $\mathbf{T}_t = [\mathbf{R}_t \mid \mathbf{t}_t] \in \mathrm{SE}(3)$ maps a world point $\mathbf{X}$ to the camera frame as $\mathbf{R}_t \mathbf{X} + \mathbf{t}_t$, with projection \mbox{$\mathbf{u} = \pi(\mathbf{K}(\mathbf{R}_t \mathbf{X} + \mathbf{t}_t))$} where $\pi([x,y,z]^\top) = [x/z, y/z]^\top$. The camera center is $\mathbf{C}_t = -\mathbf{R}_t^\top \mathbf{t}_t$. All coordinates use a metric Universal Transverse Mercator (UTM) frame.

\noindent\textbf{Map Representation.}
The reference map consists of two georeferenced rasters: (1)~the \ac{DOP} $\mathcal{D}: \mathbb{R}^2 \to \mathbb{R}^3$, an orthorectified aerial photograph providing top-down RGB at each position $(x, y)$, and (2)~the \ac{DSM} $\mathcal{S}: \mathbb{R}^2 \to \mathbb{R}$ encoding surface elevation $z = \mathcal{S}(x, y)$. Together, any DOP pixel at $(x,y)$ corresponds to the 3D point $\mathbf{X} = [\xi(x), \eta(y), \mathcal{S}(x, y)]^\top$, where $\xi$ and $\eta$ define the pixel-to-UTM affine transformation.

\noindent\textbf{Dense Feature Matching.}
We employ a pre-trained dense matcher $\mathcal{M}$ (e.g., \mbox{RoMa-v2~\cite{edstedt2025roma2}}) that, given images $\mathbf{A}$ and $\mathbf{B}$, produces $N$ correspondences\linebreak \mbox{$\{(\mathbf{u}_i^A, \mathbf{u}_i^B, c_i)\}$} with confidence $c_i \in [0,1]$. In our pipeline, $\mathbf{A}$ is the UAV frame and $\mathbf{B}$ is a DOP crop.

\noindent\textbf{Problem Formulation.}
Given video $\{\mathbf{I}_t\}$, a georeferenced map $(\mathcal{D}, \mathcal{S})$, and intrinsics $\mathbf{K}$, we aim to estimate the 6-DoF pose $\mathbf{T}_t$ for every frame in metric world coordinates, without GPS or IMU.

\subsection{First-Frame Initialization}
\label{sec:init}

OrthoTrack is fully self-initializing and requires neither GPS nor IMU at any stage. The only input beyond the video and intrinsics is a search area over the reference map, which can span the entire available \ac{DOP} coverage. We match the first UAV image against the full \ac{DOP} at reduced resolution to obtain cross-view correspondences whose DOP-side positions we convert to UTM. The median of these UTM positions serves as an approximate center around which we seed a multi-scale refinement: several \ac{DOP} crops (we use 3 crops) of increasing extent are matched independently at full resolution, and the merged 2D--3D correspondences are passed to a single PnP solve for the initial pose.

When the coarse match fails to produce a reliable position (fewer than $N_\text{gs}$ correspondences), we fall back to a coarse-to-fine pyramid tiling strategy: we partition the search area into overlapping tiles with 50\% overlap and match each tile independently. The search terminates early once a tile yields $N_\text{stop}$ PnP inliers, and its position seeds the same multi-scale refinement.

\subsection{Keyframe Localization}
\label{sec:keyframe_loc}
Each keyframe yields an absolute pose following the single-frame localization principle of OrthoLoC~\cite{dhaouadi2025ortholoc}, which matches a \ac{UAV} frame against a \ac{DOP} crop and solves PnP from \ac{DSM}-lifted correspondences. However, OrthoLoC assumes a pre-cropped \ac{DOP}/\ac{DSM} pair with high co-visibility, which is unavailable in a sequential tracking setting. We address this with a visibility-aware crop selection that handles oblique views, a two-phase matching strategy with multi-scale fallback for difficult frames, and an adaptive triggering mechanism that embeds the localizer within a tracking loop.

\noindent\textbf{Visibility-Aware Crop Selection.}
Matching the full \ac{DOP} at every keyframe is prohibitively expensive and degrades precision, since the matcher must compress a large area into a fixed-resolution input. Instead, we crop a local region around the expected camera position to preserve spatial detail.
A naive nadir-centered crop fails for oblique views where the visible ground may lie far from the point directly below the \ac{UAV}.
Given the previous pose $\mathbf{T}_{k-1}$, we project a dense grid of \ac{DSM} points into the image, retaining those that (i)~are in front of the camera, (ii)~fall within image bounds, and (iii)~are not occluded, as determined by a coarse depth buffer that discards points behind nearer surfaces.
Let $\mathcal{V}$ denote the surviving visible points with camera-frame depths $d_j$.
For oblique views ($\theta > 15^\circ$), we compute the crop center as an inverse-depth-weighted mean
\begin{equation}
    \bar{\mathbf{X}} = \frac{\sum_{j \in \mathcal{V}} d_j^{-1}\, \mathbf{X}_j}{\sum_{j \in \mathcal{V}} d_j^{-1}},
    \label{eq:vis_crop_center}
\end{equation}
which biases the crop toward the near-ground region that occupies the largest image area and where the matcher resolves detail best, rather than toward distant terrain that projects to few pixels. For near-nadir views ($\theta \leq 15^\circ$), where depth variation is small, we simply use the bounding-box center of $\mathcal{V}$. The resulting crop defines the local \ac{DOP} region $\mathcal{D}^k$ and its co-registered \ac{DSM} region $\mathcal{S}^k$, both used in the subsequent matching and lifting steps. We set the crop size proportional to the spatial extent of $\mathcal{V}$.

\noindent\textbf{Cross-View Matching and 3D Lifting.}
We match the local \ac{DOP} crop $\mathcal{D}^k$ against the UAV frame using $\mathcal{M}$, yielding correspondences $\{(\mathbf{u}_i^k, \mathbf{u}_i^{\mathcal{D}^k}, c_i)\}$. Each DOP-side match is converted to a UTM coordinate via the known georeferencing transform and lifted to 3D: $\mathbf{X}_i = [\xi(x_i), \eta(y_i), \mathcal{S}(x_i, y_i)]^\top$ using bilinear interpolation of the full-resolution DSM.
We discard correspondences with $c_i < c_{\min}$ and, if fewer than $N_{\min}$ points remain, relax the threshold to $c_{\min}'$ to retain sufficient coverage.
As a fast path, we first attempt PnP on this single crop. If it yields enough inliers, we accept the pose immediately. Otherwise, we match additional crops of increasing extent centered on the initial matches, merge the resulting correspondences, and re-run PnP.

\noindent\textbf{PnP Pose Estimation.}
We pass the resulting 2D--3D correspondences to PnP-RANSAC (SQPnP solver~\cite{terzakis2020consistently}), which jointly rejects outliers and recovers $\mathbf{T}_k$. To avoid conditioning issues with large UTM coordinates, we subtract the 3D centroid before solving and restore it in the recovered translation. Because every 3D point is sampled from the georeferenced DSM, the resulting pose is metric and globally anchored by construction.

\subsection{Inter-Frame Tracking}
\label{sec:interframe}

Once a keyframe pose is established, subsequent frames are processed online by propagating the existing 2D--3D correspondences forward via optical flow $\mathcal{O}$, avoiding the cost of repeated cross-view matching.

\noindent\textbf{Correspondence Propagation.}
Let $\mathcal{C}_k = \{(\mathbf{u}_i^{k}, \mathbf{X}_i)\}$ be the set of 2D--3D correspondences established at keyframe $k$. For each subsequent frame $t > k$, we track each 2D keypoint from $\mathbf{I}_{t-1}$ to $\mathbf{I}_t$ using $\mathcal{O}$. We apply a forward-backward consistency check and discard points whose round-trip displacement exceeds $\tau_{\text{fb}}$ pixels. Because the 3D points are anchored in the static map, the propagated pairs $\{(\mathbf{u}_i^{t}, \mathbf{X}_i)\}$ directly yield a pose via PnP-RANSAC.

\begin{figure}[ht]
    \centering
    \includegraphics[width=0.55\linewidth]{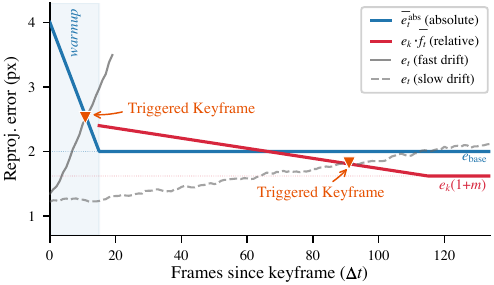}
    \caption{\textbf{Adaptive Reprojection Thresholds.} The absolute threshold (blue) starts relaxed during warmup, then settles at $e_{\text{base}}$. The relative threshold (red) tightens over $D$~frames. Rapid drift is caught by the absolute bound, slow drift by the tightening relative bound.}
    \label{fig:adaptive_threshold}
%\vspace{-0.5\baselineskip}
\end{figure}

\noindent\textbf{Keyframe Triggering.}
In principle, keyframes could be triggered at fixed intervals based on pre-defined thresholds. However, the optimal interval varies with scene content and camera motion, making a fixed schedule either wasteful during stable flight or insufficient during rapid changes. Instead, we trigger a new keyframe only when tracking quality degrades. Let $M_t$ be the number of surviving correspondences at frame~$t$, $M_k$ the count at keyframe $k$, and $\Delta t = t{-}k$ the elapsed frames. We trigger a new keyframe when any of the following conditions is met:
(1)~$M_t < N_{\min}$ (insufficient points for reliable PnP),
(2)~$M_t < \alpha \, M_k$ (large fraction of correspondences lost since the last keyframe),
(3)~$\min(\sigma_x,\sigma_y) < \sigma_{\min}$ where $\sigma_x$ and $\sigma_y$ are the spatial standard deviations of $\mathbf{u}^t$ (spatial collapse: remaining points cluster in a small image region, making PnP ill-conditioned), or
(4)~reprojection error $e_t$ exceeds an adaptive threshold.

A fixed reprojection threshold for condition~(4) is problematic: cross-view matching produces a variable initial error at each keyframe, so a tight threshold triggers an infinite re-triggering loop, while a loose one misses slow drift. We therefore combine two adaptive thresholds that adjust to each keyframe's matching quality (\cref{fig:adaptive_threshold}), making the criterion robust across scenes without per-dataset tuning.
The first is an absolute threshold that relaxes the bound during a warmup window after each keyframe:
\begin{equation}
    \bar{e}_t^{\,\text{abs}} = e_{\text{base}} + \delta \cdot \max\!\bigl(0,\; 1 - \tfrac{\Delta t}{G}\bigr),
    \label{eq:reproj_abs}
\end{equation}
where $e_{\text{base}}$ is the steady-state bound, $\delta$ the initial tolerance, and $G$ the warmup duration in frames. The second is a relative threshold, active for $\Delta t \geq G$, that detects slow drift by comparing $e_t$ to the keyframe's baseline error $e_k$. It triggers when $e_t > \bar{f}_t \cdot e_k$, where
\begin{equation}
    \bar{f}_t = 1 + (f_0 {-} 1)(1 {-} \rho) + m\,\rho, \quad \rho = \min\!\bigl(1,\; \tfrac{\Delta t - G}{D}\bigr),
    \label{eq:reproj_growth}
\end{equation}
decreases from $f_0$ to $1{+}m$ over $D$ frames. Since this threshold scales with $e_k$, it adapts to each keyframe's matching quality: a noisier keyframe permits more absolute error, while the decaying factor ensures gradual drift eventually triggers re-localization.

\section{The MovingDrone Dataset}
\label{sec:dataset}
\begin{figure*}[t]
    \centering
    \includegraphics[width=0.92\linewidth]{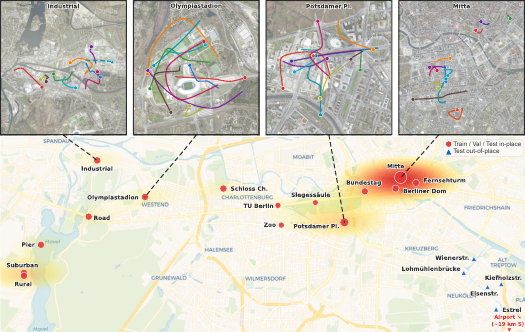}
    \caption{\textbf{MovingDrone Locations.}
        194 sequences across 21 Berlin regions. Insets show representative trajectories over satellite imagery. The heatmap encodes sequence density.}
    \label{fig:uavseq_locations}
%\vspace{-0.5\baselineskip}
\end{figure*}
We present MovingDrone, a large-scale UAV localization benchmark that renders photorealistic video from the Berlin VirtualCityMap~\cite{virtualcitymap_berlin}, a high-precision textured photogrammetric mesh reconstructed from real aerial imagery (cross-validated to 0.18\,m median vertical accuracy) that covers Germany's largest city.
We select 21 regions to span Berlin's full architectural and landscape diversity, from dense historical centers and modern high-rises to parks, waterways, airports, and stadiums.
The fully automated pipeline can generate arbitrarily many sequences anywhere within the city's coverage at negligible cost, circumventing the airspace restrictions, privacy regulations, and expensive RTK-GNSS supervision that make real UAV data collection impractical at metropolitan scale.
Because the mesh preserves authentic textures, lighting, and geometric detail from the original survey, the synthetic-to-real domain gap is substantially smaller than in game-engine datasets such as GTA-UAV~\cite{gta_uav_25}. The underlying reconstruction stems from a comprehensive campaign, yielding a watertight, gap-free surface, unlike meshes in other benchmarks~\cite{wu2023uavd4l,wang2025uavscenes} that suffer from holes in areas due to insufficient image coverage.
Each sequence is paired with the same publicly available geodata that a practitioner would use in the field, ensuring that every label is by construction spatially consistent with operational map assets at sub-pixel accuracy.

\begin{figure*}[t]
    \centering
    \includegraphics[width=\linewidth]{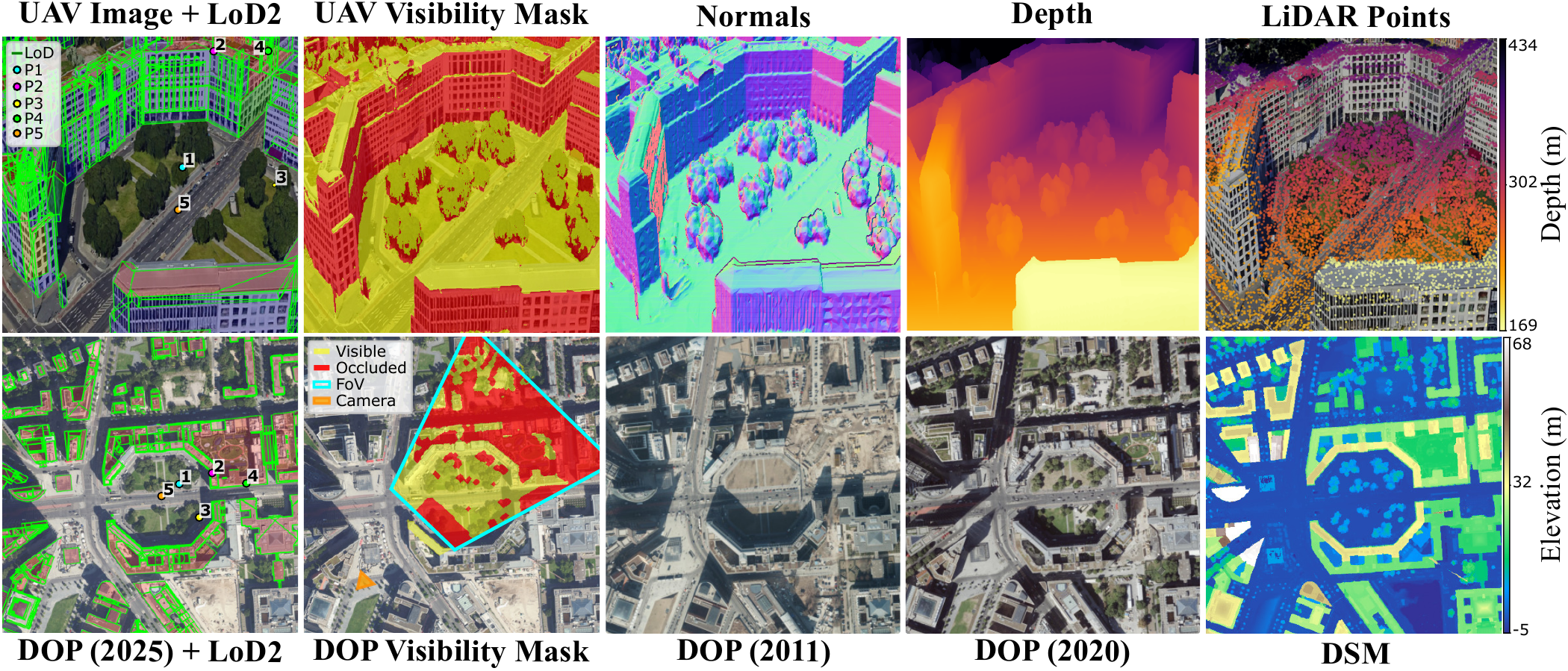}
    \caption{\textbf{MovingDrone Multi-Modal Data Per Frame.}
    Each rendered UAV frame is paired with depth, surface normals from the mesh, LiDAR, multi-year DOPs, a DSM, and LoD1/2 building meshes. Comparing the three DOPs reveals the temporal domain gap: lighting, vegetation, and even building footprints change across years.}
    \label{fig:dataset_samples}
%\vspace{0\baselineskip}
\end{figure*}

\noindent \textbf{Scale and Diversity.}
MovingDrone comprises 194 sequences totaling 343{,}114 frames across 21 regions (\cref{fig:uavseq_locations}).
Camera tilt ranges from near-nadir ($0^\circ$) to highly oblique ($83^\circ$), with 92\% of frames exceeding $20^\circ$, reflecting the oblique-view regime common in inspection and monitoring applications.
On average, each sequence is paired with 13.19 DOP years (2011--2025), enabling analysis of localization robustness to map age (\cref{fig:dataset_samples}).

\noindent \textbf{Generation Pipeline.}
Camera trajectories are designed in Google Earth Studio across 21 Berlin regions covering four archetypes (orbital, linear, oblique descent, multi-turn) at 64--1{,}183\,m altitude, inspired by the trajectories of public drone footages.
The exported trajectories are resampled via cubic splines to bound speeds to 10--100\,km/h.
Although the base trajectories are designed rather than recorded from a physical drone, three physically motivated augmentations inject realistic flight dynamics: (i)~multi-band trajectory noise combining GPS drift with motor vibration, (ii)~stochastic wind-gust events, and (iii)~velocity-dependent directional motion blur.
Photorealistic video is rendered from the Berlin VirtualCityMap~\cite{virtualcitymap_berlin} textured mesh, with per-frame depth obtained through ray casting for pixel-exact consistency.
Each sequence is automatically paired with open geodata from Berlin and Brandenburg portals~\cite{geoportal_berlin,geoportal_brandenburg} (\cref{fig:dataset_samples}): multi-year DOPs and a DSM (both at 0.2\,m resolution), classified airborne LiDAR (5--20\,pts/m$^2$, ASPRS classes), the textured photogrammetric mesh used for rendering, and CityGML building meshes (LoD1/LoD2).
Full details on the generation pipeline, augmentations, and data splits are provided in the supplement.

\section{Experiments}
\label{sec:experiments}
\begin{table*}[ht]
\centering
\caption{\textbf{6-DoF UAV Trajectory Estimation on UAVScenes and MovingDrone.} \textbf{Bold}: best; \underline{underline}: second best. Metrics: ATE and TE in meters [m]$\downarrow$, RE in degrees [$^\circ$]$\downarrow$, R@$k$ in percent [\%]$\uparrow$.}
\vspace{-0.5\baselineskip}
\label{tab:main_comparison}
\setlength{\tabcolsep}{1.5pt}
\renewcommand{\arraystretch}{0.88}
\scriptsize
\resizebox{\textwidth}{!}{%
\begin{tabular}{l@{\hspace{2pt}}l@{\hspace{3pt}}rrrrr@{\hspace{5pt}}rrrrrr}
\toprule
\multirow{2}{*}{Method} & \multirow{2}{*}{Align.} & \multicolumn{5}{c}{UAVScenes~\cite{wang2025uavscenes}} & \multicolumn{6}{c}{MovingDrone (Ours)} \\
\cmidrule(lr){3-7} \cmidrule(lr){8-13}
& & ATE$\downarrow$ & TE$\downarrow$ & RE$\downarrow$ & R@1$\uparrow$ & R@2$\uparrow$ & ATE$\downarrow$ & TE$\downarrow$ & RE$\downarrow$ & R@1$\uparrow$ & R@2$\uparrow$ & FPS$\uparrow$ \\
\midrule
\multicolumn{13}{l}{\emph{VO / SLAM}} \\
Five-Point VO~\cite{nister2004efficient} & Sim(3) & 76.08 & 61.98 & 28.53 & 0.0 & 0.2 & 55.89 & 45.25 & 36.76 & 0.0 & 0.0 & 2.0 \\
ORB-SLAM3~\cite{orbslam3} & ff+s & 84.39 & 56.48 & 29.25 & \textbf{5.6} & 8.5 & 29.20 & 25.45 & 10.25 & 12.1 & 27.6 & \textbf{22.3} \\
ORB-SLAM3~\cite{orbslam3} & Sim(3) & 36.90 & 20.58 & \underline{14.61} & 1.3 & 6.3 & 13.14 & 9.11 & 14.30 & 5.9 & 22.9 & \textbf{22.3} \\
ORB-SLAM3~\cite{orbslam3} & G, pw & \underline{6.98} & \underline{1.47} & 108.83 & 0.6 & 3.6 & 5.24 & 2.37 & 48.90 & 0.8 & 5.7 & \textbf{22.3} \\
ORB-SLAM3~\cite{orbslam3} & G+M, pw & \underline{6.98} & \underline{1.47} & 99.47 & 2.7 & 8.7 & 5.24 & 2.37 & 42.82 & 0.8 & 7.6 & \textbf{22.3} \\
DROID-SLAM~\cite{teed2021droid} & ff+s & 147.52 & 126.52 & 51.49 & 3.4 & 4.4 & 0.89 & 0.75 & \underline{0.14} & 72.7 & 89.6 & \underline{4.7} \\
DROID-SLAM~\cite{teed2021droid} & Sim(3) & 103.96 & 86.81 & 59.85 & 0.5 & 3.0 & \underline{0.34} & \underline{0.22} & 0.47 & \underline{88.8} & \underline{89.7} & \underline{4.7} \\
DROID-SLAM~\cite{teed2021droid} & pw & 8.03 & 2.53 & \textbf{0.80} & \underline{5.4} & \textbf{32.1} & 0.63 & 0.35 & \textbf{0.08} & \textbf{93.7} & \textbf{99.1} & \underline{4.7} \\
DROID-SLAM~\cite{teed2021droid} & G, pw & \textbf{1.04} & \textbf{0.08} & 46.63 & 1.7 & 6.1 & \textbf{0.13} & \textbf{0.05} & 4.57 & 24.1 & 37.5 & \underline{4.7} \\
DROID-SLAM~\cite{teed2021droid} & G+M, pw & \textbf{1.04} & \textbf{0.08} & 35.02 & 5.0 & \underline{12.2} & \textbf{0.13} & \textbf{0.05} & 4.30 & 25.7 & 41.2 & \underline{4.7} \\
DPVO~\cite{teed2024deep} & ff+s & 151.06 & 128.21 & 49.78 & 3.1 & 3.5 & 4.01 & 2.06 & 0.37 & 63.8 & 82.2 & 4.5 \\
DPVO~\cite{teed2024deep} & Sim(3) & 100.92 & 85.78 & 53.88 & 0.1 & 0.9 & 2.43 & 1.04 & 0.68 & 72.7 & 84.1 & 4.5 \\
\midrule
\multicolumn{13}{l}{\emph{Foundation models}} \\
DUSt3R~\cite{wang2024dust3r}$^\dagger$ & Sim(3) & 84.14 & 63.37 & 37.84 & \underline{0.0} & \underline{0.0} & 53.43 & 43.60 & 81.32 & 0.0 & 0.0 & 0.2 \\
DA3-Nested~\cite{lin2025depth} & Sim(3) & \underline{58.27} & \underline{38.74} & \underline{16.47} & \textbf{0.2} & \textbf{1.1} & \textbf{2.74} & \textbf{1.83} & \textbf{2.57} & \textbf{28.7} & \textbf{51.4} & \underline{11.3} \\
VGGT-SLAM v2~\cite{wang2025vggt} & Sim(3) & \textbf{57.17} & \textbf{34.57} & \textbf{9.47} & \underline{0.0} & \underline{0.0} & \underline{11.94} & \underline{9.28} & \underline{15.88} & 0.0 & 0.0 & 3.1 \\
Pi3~\cite{wang2025pi3}$\ddagger$ & Sim(3) & 102.80 & 80.39 & 53.75 & \underline{0.0} & \underline{0.0} & 23.98 & 19.30 & 39.46 & \underline{0.2} & \underline{14.2} & \textbf{25.2} \\
\midrule
\multicolumn{13}{l}{\emph{UAV localization (GPS/IMU prior)}} \\
LoD-Loc~\cite{zhu2024lodloc} & none & \multicolumn{5}{c}{\textit{no LoD models}} & 17.12 & 9.52 & 1.49 & 4.0 & 11.5 & 3.0 \\
LoD-Loc~\cite{zhu2024lodloc}~(retrained) & none & \multicolumn{5}{c}{\textit{no LoD models}} & 10.04 & 5.25 & 1.10 & 10.8 & 30.3 & 3.3 \\
OrthoLoC~\cite{dhaouadi2025ortholoc}$^*$ & none & 38.24 & \textbf{1.26} & \textbf{0.43} & \textbf{23.5} & \textbf{93.8} & 19.78 & 0.50 & \underline{0.08} & 85.4 & 95.4 & 0.5 \\
\midrule
OrthoTrack -- loc.\ only & none & \underline{1.89} & \underline{1.28} & \textbf{0.43} & 20.5 & \underline{91.7} & \underline{4.83} & \textbf{0.30} & \textbf{0.06} & \textbf{95.8} & \textbf{99.2} & 1.9 \\
OrthoTrack -- track only & none & 102.64 & 19.53 & 7.75 & 12.3 & 15.0 & 27.70 & 16.36 & 5.20 & 3.0 & 7.7 & \underline{13.4} \\
OrthoTrack (Ours) & none & \textbf{1.51} & 1.38 & \underline{0.49} & \underline{20.6} & 88.0 & \textbf{0.67} & \underline{0.33} & \textbf{0.06} & \underline{90.9} & \underline{97.9} & \textbf{23.8} \\
\bottomrule
\end{tabular}}%\par
\vspace{0.25\baselineskip}
{\raggedright\scriptsize Align.: Sim(3) = oracle 7-DoF, ff+s = first-frame anchor + scale, pw = piecewise Sim(3) to geodata anchors, none = absolute poses (no alignment). $^\dagger$\,windowed; $\ddagger$\,sometimes windowed; $^*$\,GIM(DKM)+AdHoP, same DOP/DSM inputs as OrthoTrack.\par}
\end{table*}

\sloppy

In the following, we evaluate OrthoTrack's trajectory estimation accuracy on MovingDrone and the real-world UAVScenes~\cite{wang2025uavscenes} benchmark against \ac{VO}/\ac{SLAM} systems, 3D foundation models, and geodata-based localization methods, and validate each design choice through ablation studies. 

\subsection{Experimental Setup}
\label{sec:exp_setup}

\noindent\textbf{Datasets.}
We evaluate on the 10 MovingDrone test sequences (with \ac{DOP} from year 2025) and all 20 real-world UAVScenes~\cite{wang2025uavscenes} sequences (121\,k real-world frames at 10\,fps) across Armenia and Hong Kong.
Since UAVScenes ships neither \acp{DOP} nor \acp{DSM}, we render both from the provided \ac{LiDAR}-reconstructed mesh at 0.10\,m GSD. In contrast, MovingDrone uses independently surveyed geodata, producing a genuine cross-domain gap, while the cross-domain gap in UAVScenes is small.
No dataset-specific adaptation is applied.

\noindent\textbf{Metrics.}
We report six metrics, all computed per sequence and averaged: ATE (RMSE of 3D position errors), TE (median translation error), RE (median rotation error), and joint recall R@$k$ (percentage of frames with TE${<}k$\,m and RE${<}k^\circ$). All timing is measured on a single NVIDIA L40S GPU.

\noindent\textbf{Baselines.}
We compare in~\cref{tab:main_comparison} against monocular \ac{VO}/\ac{SLAM} systems, recent 3D foundation models, and geodata-based localization methods.
All baselines and models in the ablation studies use official code with default hyperparameters.

\noindent\textbf{Evaluation Protocol.}
\ac{VO}/\ac{SLAM} systems receive raw video under oracle Sim(3) alignment and first-frame anchoring with scale correction (ff+s).
Foundation models not fitting to GPU memory use sliding-window inference with per-window Sim(3) stitching.
OrthoLoC~\cite{dhaouadi2025ortholoc} and LoD-Loc~\cite{zhu2024lodloc} receive simulated sensor priors (noisy GPS + orientation). Piecewise-aligned DROID-SLAM (pw) aligns a pre-computed trajectory every 50 frames via Sim(3) using the same matcher and pose solver as OrthoTrack keyframes.

\noindent\textbf{Implementation Details.}
We use RoMa-v2~\cite{edstedt2025roma2} Precise as the dense matcher, selected for its strong cross-domain performance, and GPU-accelerated Lucas-Kanade~\cite{lucas1981iterative} for inter-frame optical flow. The design choice is justified in~\cref{tab:ablation}.
The PnP-RANSAC pose solver uses SQPnP~\cite{terzakis2020consistently}. The remaining hyperparameters and timing details are in the supplementary.

\begin{figure}[t]
    \centering
    \includegraphics[width=\linewidth]{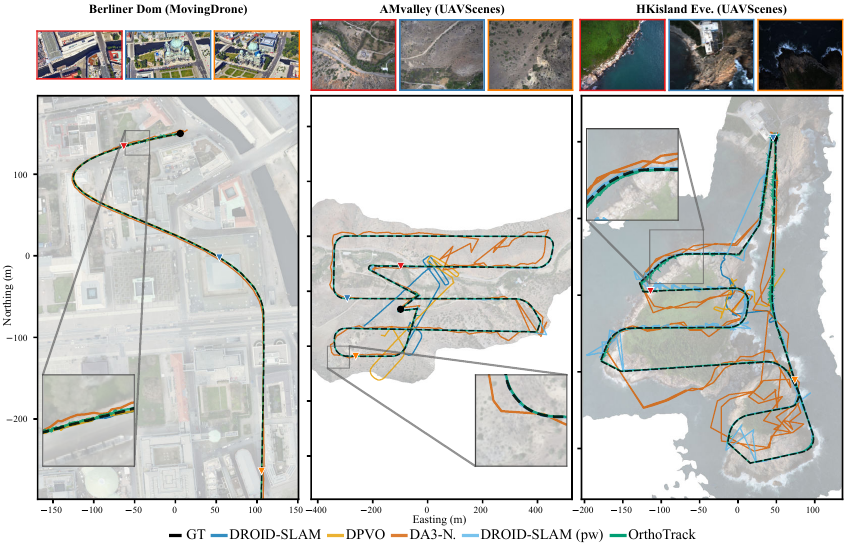}
    \caption{\textbf{Trajectory Comparison.} VO/SLAM baselines receive oracle Sim(3) alignment yet drift or collapse on long flights. Piecewise-anchored DROID-SLAM (pw) reduces drift but accumulates errors between anchors. OrthoTrack produces absolute poses without alignment and remains accurate.}
    \label{fig:trajectory_comparison}
%\vspace{-0.5\baselineskip}
\end{figure}
\subsection{Comparison with State of the Art}
\label{sec:comparison}

\begin{figure}[ht]
    \centering
    \includegraphics[width=0.93\linewidth]{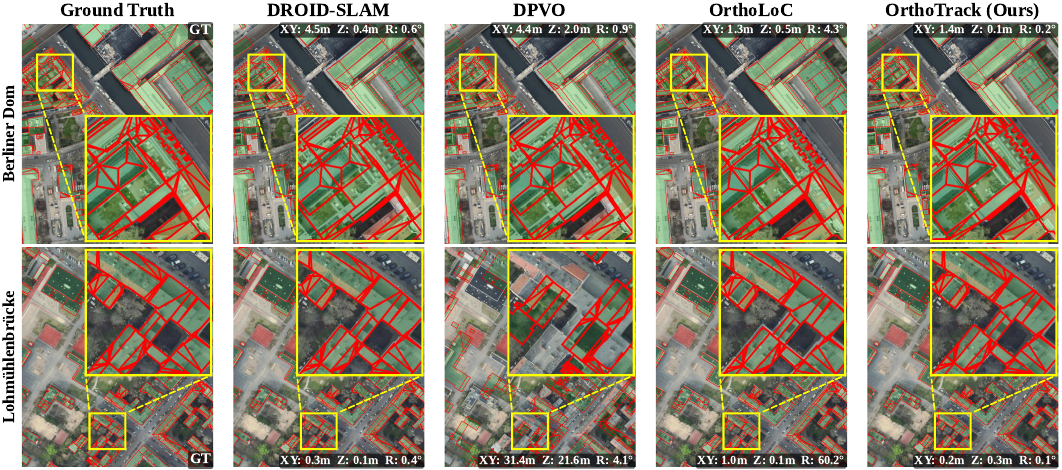}
    \caption{\textbf{Per-Frame Localization Quality.} \ac{LoD}2 building wireframes (red) overlaid on the orthophoto are shifted by each method's position error, making misalignment directly visible.}
    \label{fig:localization_quality}
%\vspace{-0.5\baselineskip}
\end{figure}

\cref{tab:main_comparison} presents results on MovingDrone and on real-world UAVScenes~\cite{wang2025uavscenes}.
OrthoTrack is the only method to maintain sub-metre ATE on both benchmarks without oracle alignment or sensor priors, while operating at real-time speed.
Representative trajectories are shown in \cref{fig:trajectory_comparison} and per-frame localization quality is visualized in \cref{fig:localization_quality}.

Among \ac{VO}/\ac{SLAM} methods, DROID-SLAM~\cite{teed2021droid} achieves the lowest ATE on MovingDrone under oracle Sim(3) (0.34\,m), yet collapses to 104\,m on UAVScenes (a $69{\times}$ gap to OrthoTrack).
We attribute this to the near-planar scene geometry at high altitude: monocular parallax cannot resolve depth when the baseline-to-depth ratio is negligible, causing the reconstructed trajectory to compress and accumulate drift that no post-hoc alignment can recover.
Piecewise geodata anchoring (pw) even slightly outperforms OrthoTrack on MovingDrone (0.63 vs.\ 0.67\,m ATE). On UAVScenes, however, the VO collapse propagates between anchors (8.03 vs.\ 1.51\,m, ${>}5{\times}$ above OrthoTrack).
Providing piecewise alignment with oracle \ac{GNSS} (G) and magnetometer (M) significantly improves the \ac{VO} baselines by continuously re-anchoring the scale and heading, allowing DROID-SLAM to reach 1.04\,m ATE on UAVScenes. However, the addition of the magnetometer (+M) does not significantly reduce the overall rotation error (RE). This is because a magnetometer provides only a global yaw reference and cannot correct accumulated drift in pitch and roll, leaving the 3D orientation degraded. Furthermore, this piecewise approach relies on perfect continuous sensor priors, which are often unavailable or unreliable in real-world aerial scenarios.
DPVO~\cite{teed2024deep} and ORB-SLAM3~\cite{orbslam3} exhibit the same degeneracy, confirming a fundamental limitation of relative-pose methods on aerial scenes. Moreover, DROID runs at 4.7\,FPS vs.\ OrthoTrack's 23.8\,FPS, making it $5{\times}$ slower even before accounting for the alignment oracle it requires.

Foundation models fare similarly: despite oracle alignment, even the strongest variant cannot approach OrthoTrack's performance. This gap is primarily driven by out-of-distribution bias, as these models were trained predominately on ground-level images and struggle with the aerial perspective. Furthermore, because they predict geometry in a local frame, they rely on a similarity transform that breaks down when the reconstruction drifts over long sequences without any mechanism for external re-anchoring.

Among geodata-based methods that use prior sensor data (GPS/IMU) to identify the region of interest, LoD-Loc~\cite{zhu2024lodloc} reaches only 10\,m ATE on MovingDrone even after retraining, likely because wireframe alignment is less discriminative than dense texture matching on complex urban scenes or due to degenerate results in regions with limited buildings.
OrthoLoC~\cite{dhaouadi2025ortholoc} achieves competitive single-frame TE (0.50 vs.\ 0.33\,m on MovingDrone and 1.26 vs.\ 1.38\,m on UAVScenes), confirming per-frame localizer accuracy, yet without temporal consistency sporadic failures propagate directly into ATE (19.78\,m and 38.24\,m).
OrthoTrack cuts ATE by $30{\times}$ and by $25{\times}$ on the respective datasets, invoking the matcher at only ${\sim}1\%$ of frames and propagating correspondences via optical flow, effectively filtering isolated outliers without GPS or IMU.

\begin{table*}[ht]
\centering
\caption{\textbf{Ablation Studies on MovingDrone.} Each section varies one pipeline component while fixing all others. \textbf{Bold}: best; \underline{underline}: second best (per section).}
\vspace{-0.5\baselineskip}
\label{tab:ablation}
\setlength{\tabcolsep}{3.5pt}
\renewcommand{\arraystretch}{0.7}
\scriptsize
\begin{tabular}{l@{\hspace{5pt}}rrrrrr@{\hspace{4pt}}r@{\hspace{5pt}}rr}
\toprule
Config & ATE$\downarrow$ & TE$\downarrow$ & RE$\downarrow$ & R@1$\uparrow$ & R@2$\uparrow$ & R@5$\uparrow$ & \#KF & FPS$_\text{c}$$\uparrow$ & FPS$_\text{s}$$\uparrow$ \\
\midrule
\multicolumn{10}{l}{\emph{Flow backend (matcher = RoMa v2 Precise, 30\,fps)}} \\
LK~\cite{lucas1981iterative} & \underline{0.67} & \textbf{0.33} & \textbf{0.06} & 90.9 & \underline{97.9} & 99.3 & 15 & \textbf{30.3} & \textbf{23.8} \\
RAFT~\cite{teed2020raft} & \textbf{0.66} & \underline{0.37} & \underline{0.08} & 91.9 & \textbf{98.4} & \textbf{99.6} & \textbf{10} & 11.9 & 10.0 \\
SEA-RAFT~\cite{wang2024sea} & 1.93 & \underline{0.37} & \underline{0.08} & \textbf{93.5} & \underline{97.9} & \underline{99.4} & 12 & 17.4 & 14.1 \\
NeuFlow v2~\cite{zhang2025neuflow} & 0.82 & \textbf{0.33} & \textbf{0.06} & \underline{92.9} & \underline{97.9} & 98.7 & \underline{11} & \underline{24.4} & \underline{17.6} \\
\midrule
\multicolumn{10}{l}{\emph{Keyframe matcher (flow = LK, 30\,fps)}} \\
RoMa v2 (Precise) & \textbf{0.67} & \underline{0.33} & \textbf{0.06} & \underline{90.9} & \textbf{97.9} & \underline{99.3} & 15 & 1.3 & 23.8 \\
RoMa v2 (Base) & 0.90 & 0.49 & \underline{0.10} & 86.4 & 95.9 & 99.0 & \underline{13} & \underline{2.3} & \textbf{26.3} \\
RoMa v2 (Fast) & \underline{0.88} & 0.58 & 0.11 & 79.9 & \underline{96.9} & 99.1 & \underline{13} & \textbf{2.6} & \underline{25.4} \\
GIM(DKM)~\cite{wang2024efficient} & 2.56 & 0.52 & \underline{0.10} & 89.1 & \textbf{97.9} & \textbf{99.4} & \underline{13} & 0.7 & 21.7 \\
L2M~\cite{liang2025learning} & 6.23 & \textbf{0.29} & \textbf{0.06} & \textbf{94.9} & 96.4 & 98.3 & 15 & 0.4 & 15.0 \\
MASt3R~\cite{leroy2024grounding} & 5.74 & 2.10 & 0.46 & 20.4 & 56.3 & 85.6 & \textbf{11} & 0.3 & 5.5 \\
\midrule
\multicolumn{10}{l}{\emph{Input frame rate (matcher = RoMa v2 Precise, flow = LK)}} \\
30\,fps (1$\times$) & \underline{0.67} & \textbf{0.33} & \textbf{0.06} & \underline{90.9} & \textbf{97.9} & 99.3 & 15 & -- & \textbf{23.8} \\
10\,fps (3$\times$) & \textbf{0.58} & 0.39 & 0.08 & \textbf{91.0} & \textbf{97.9} & \textbf{99.7} & \underline{11} & -- & \underline{19.2} \\
3\,fps (10$\times$) & 0.68 & 0.42 & 0.08 & 90.6 & \underline{97.8} & 99.0 & \textbf{8} & -- & 11.1 \\
2\,fps (15$\times$) & 0.71 & \underline{0.38} & \underline{0.07} & \underline{90.9} & 97.3 & \underline{99.4} & \textbf{8} & -- & 8.7 \\
\midrule
\multicolumn{10}{l}{\emph{Keyframe triggering (matcher = RoMa v2 Precise, flow = LK, 30\,fps)}} \\
Adaptive (Ours) & \underline{0.67} & \textbf{0.33} & \textbf{0.06} & \underline{90.9} & \underline{97.9} & \underline{99.3} & 15 & -- & \underline{23.8} \\
Fixed reproj ($e{=}2$\,px) & 0.79 & 0.46 & 0.08 & 85.2 & 94.9 & \textbf{99.7} & 9 & -- & 21.4 \\
Fixed interval ($K{=}50$) & \textbf{0.53} & \underline{0.35} & \textbf{0.06} & \textbf{93.3} & \textbf{99.3} & \textbf{99.7} & 28 & -- & 19.0 \\
Fixed interval ($K{=}50$, 2\,fps) & 12.26 & 1.30 & \underline{0.07} & 54.3 & 71.0 & 85.8 & \textbf{3} & -- & 6.8 \\
Fixed interval ($K{=}150$) & 0.82 & 0.41 & \textbf{0.06} & 90.7 & 96.9 & \underline{99.3} & 10 & -- & \textbf{24.2} \\
Point-count only & 1.69 & 0.93 & \textbf{0.06} & 61.1 & 80.8 & 96.0 & \underline{4} & -- & 22.5 \\
\bottomrule
\end{tabular}\par
\vspace{0.25\baselineskip}
{\raggedright\scriptsize FPS$_\text{c}$: isolated throughput of the varied component (flow tracking or matching rate).\\ FPS$_\text{s}$: end-to-end system throughput.\par}
%\vspace{-0.5\baselineskip}
\end{table*}

\subsection{Ablation Studies}
\label{sec:ablation}

\noindent\textbf{Localization vs.\ Tracking.}
The ablation variants in \cref{tab:main_comparison} isolate each component.
Localizing (loc. only) every frame via matching, similar to OrthoLoC but with RomaV2 and visibility-aware crop selection and without AdHoP strategy~\cite{dhaouadi2025ortholoc}, achieves the lowest TE but at prohibitive cost (1.9\,FPS). Tracking alone (track only) by initializing the first keyframe and iteratively creating a new set tracking points at each frame without global re-anchoring drifts.
The full pipeline combines both: flow-based tracking suppresses per-frame outliers while periodic keyframe localization corrects accumulated drift.
On UAVScenes, loc-only achieves lower TE (1.28 vs.\ 1.38\,m) but higher ATE (1.89 vs.\ 1.51\,m), confirming that trajectory consistency requires temporal filtering beyond per-frame accuracy.

\noindent\textbf{Optical Flow Backend.}
\cref{tab:ablation} compares four flow backends.
LK~\cite{lucas1981iterative}, RAFT~\cite{teed2020raft}, and NeuFlow v2~\cite{zhang2025neuflow} achieve sub-metre ATE.
We attribute LK's parity with learned methods to the pipeline's regime: at 30\,fps, inter-frame displacements are small enough for a classical pyramid tracker, and keyframe resets bound drift before it affects accuracy. We select LK for its best accuracy-speed trade-off.

\noindent\textbf{Matching Method.}
\cref{tab:ablation} compares matchers spanning dense and standalone methods.
RoMa-v2~\cite{edstedt2025roma2} Precise achieves the best ATE (0.67\,m), which we attribute to its training on AerialMegaDepth~\cite{vuong2025aerialmegadepth} that targets the air-to-ground domain gap.
Five of six matchers produce sub-metre TE, confirming that the pipeline could be designed with any matcher. The end-to-end throughput stays similar because keyframes constitute only ${\sim}1\%$ of frames.

\noindent\textbf{Input Frame Rate.}
\cref{tab:ablation} evaluates robustness from 30 down to 2\,fps. Accuracy remains stable because the adaptive monitor self-regulates: at lower rates it triggers re-localization more often (${\sim}0.6\%$ at 30\,fps to ${\sim}5\%$ at 2\,fps), automatically trading throughput for accuracy.

\noindent\textbf{Keyframe Triggering Strategy.}
\cref{tab:ablation} compares the adaptive dual-threshold against simpler heuristics.
Fixing a threshold for the reprojection error does not outperform our adaptive strategy since the reprojection error alone does not sufficiently reflect the quality of the estimated pose. Using a regular interval of keyframes show different results: For small intervals ($K=50$), the results are slightly better than our adaptive strategy as keyframes are triggered more often. However, this degradates with lower frame rate (2\,~fps instead of 30\,~fps), indicating the sensitivity of this parameter to the camera motion. Yet, our adaptive strategy captures both early and slow drifts: the grace ramp prevents re-triggering loops when cross-view reprojection error is initially high, while the decaying relative bound catches slow geometric drift.

Further evaluation on UAVD4L~\cite{wu2023uavd4l}, additional analysis and details are provided in the supplementary material.

\section{Conclusion}
\label{sec:conclusion}

We present OrthoTrack, a training-free pipeline that matches UAV frames against publicly available orthophotos and surface models, yielding drift-free, metrically scaled 6-DoF poses without GPS or training.
We also introduce MovingDrone, a multi-modal benchmark of 194 sequences across 21 Berlin regions whose scalable generation pipeline and paired geodata support broader aerial perception research.
On MovingDrone, OrthoTrack achieves 0.67\,m ATE and 0.33\,m median translation error without any alignment or sensor prior. On the external real-world UAVScenes benchmark it reduces ATE by $69{\times}$ over the best sim(3)-aligned VO system.
Unlike VO/SLAM systems that drift or foundation models bound to a local frame, OrthoTrack anchors every pose in public geodata, enabling deployment to new regions without site-specific data.

\vspace{1em}
\noindent\textbf{Acknowledgments.} This research was partially funded by the ETH Foundation Project 2025-FS-352, the SNSF Advanced Grant 216260.

% ---- Bibliography (excluded from 14-page limit) ----
\bibliographystyle{splncs04}
\bibliography{main}

% ---- Supplementary Material ----
\clearpage
\appendix
\renewcommand{\thesection}{\Alph{section}}
\setcounter{section}{0}
\setcounter{figure}{0}
\setcounter{table}{0}
\renewcommand{\thefigure}{\Alph{section}\arabic{figure}}
\renewcommand{\thetable}{\Alph{section}\arabic{table}}

\begin{center}
\Large\textbf{Supplementary Material}\\[0.3em]
\large\textit{OrthoTrack: Continuous 6-DoF UAV Trajectory Estimation Anchored in Public Orthophotos}\\[0.5em]
\end{center}
\vspace{1em}

\noindent\textbf{Overview.}
\Cref{sec:supp_protocol} covers implementation details, runtime and resource usage, and hyperparameter sensitivity.
\Cref{sec:supp_dataset} describes the MovingDrone generation pipeline, realism augmentations, dataset splits and statistics, dataset visualization, and licensing.
\Cref{sec:supp_experiments} presents additional experiments: foundation model evaluation under two alignment protocols, evaluation on UAVD4L, analysis of challenging scenarios, sensor prior noise sensitivity, geodata sensitivity, extended matcher ablation, \ac{DOP} year analysis, and MovingDrone as a monocular depth benchmark.
\Cref{sec:supp_future} discusses limitations and future work.
\Cref{sec:supp_ethics} discusses ethical considerations and broader impact.

\section{Pipeline Details}
\label{sec:supp_protocol}

\subsection{Implementation Details}
We use RoMa-v2~\cite{edstedt2025roma2} in precise mode as the dense matcher $\mathcal{M}$, sampling up to $N = 3{,}000$ correspondences with confidence threshold $c_{\min} = 0.4$ (relaxed to $0.1$ when too few matches pass), selected for its strong cross-domain performance (\cref{tab:supp_matcher_full}).
Inter-frame optical flow uses Lucas-Kanade with a $21{\times}21$ window, 3 pyramid levels, and forward-backward threshold $\tau_{\text{fb}} = 1.0$\,px, providing the best accuracy-speed trade-off.
PnP-RANSAC uses SQPnP~\cite{terzakis2020consistently} with reprojection thresholds of 4.0\,px for keyframes and 7.0\,px for tracked frames. A crop is accepted when its inlier ratio exceeds $r_{\min} = 0.30$.
A new keyframe is triggered when tracked points fall below $N_{\min} = 100$ or the point-drop ratio exceeds $\alpha = 0.40$. The spatial-collapse radius $\sigma_{\min} = 30$\,px prevents triggering when tracked points cluster in a small image region.
The adaptive reprojection threshold starts at $e_{\text{base}} = 2.0$\,px and grows by $\delta = 2.0$\,px per consecutive failure, with grace ramp $G = 15$ frames, growth factor $f_0 = 2.0$, growth margin $m = 0.35$, and decay period $D = 100$ frames.
Visibility-aware cropping samples the \ac{DSM} on a 5\,m grid within 800\,m with 16\,px z-buffer cells.
First-frame initialization performs full-\ac{DOP} matching at reduced resolution. If this fails, a coarse-to-fine pyramid fallback dynamically subdivides the \ac{DOP}, stopping early as soon as any tile yields a reliable pose with at least 30 inliers.
All hyperparameters were set once and used unchanged across all three evaluation datasets (MovingDrone, UAVScenes~\cite{wang2025uavscenes}, UAVD4L~\cite{wu2023uavd4l}) without per-dataset tuning.

\subsection{Runtime and Resource Usage}
\label{sec:supp_timing}

\cref{tab:timing} reports per-component latencies averaged over 10 test sequences.
The bottleneck is keyframe localization (RoMa-v2 matching + multi-crop PnP) at 1.4\,s per event, amortized over inter-frame optical flow steps at 23\,ms each, yielding 23.8\,FPS and comfortably exceeding real-time for 30\,FPS video.
Faster RoMa-v2 variants (Base/Fast) reduce keyframe cost to 0.4--0.5\,s while preserving submeter accuracy (see ablations in main paper), though overall throughput is dominated by inter-frame flow steps, so the end-to-end speedup is modest.
Peak GPU memory is approximately 11\,GB during keyframe events, driven by the RoMa-v2 Precise forward pass at $1{,}280{\times}1{,}280$ resolution. Between keyframes, only ${\sim}$2.1\,GB is needed for Lucas-Kanade flow and PnP, leaving ample headroom for onboard processes. The full pipeline fits on a single consumer GPU with ${\geq}$12\,GB VRAM (e.g.\ RTX 3060/4070).
CPU RAM ranges from 6 to 18\,GB depending on geodata tile extent (mean 11\,GB).
First-frame initialization takes 3.5\,s on average (max 10.8\,s) when the search spans local \ac{DOP} coverage. A city-scale search scales linearly with tile count, but in practice the last known GPS fix constrains the area, making initialization near-instantaneous.

\begin{table}[ht]
\centering
\caption{\textbf{Per-Component Timing.} GPU: averaged over 10 test sequences on one NVIDIA L40S. CPU-only: Intel Xeon Gold 6248R.}
\label{tab:timing}
\setlength{\tabcolsep}{4pt}
\renewcommand{\arraystretch}{0.9}
\scriptsize
\begin{tabular}{@{}l rrr@{}}
\toprule
 & \multicolumn{2}{c}{RoMa: GPU} & RoMa: CPU \\
\cmidrule(lr){2-3} \cmidrule(lr){4-4}
Component & LK: GPU & LK: CPU & LK: CPU \\
\midrule
First-frame initialization & 3.5\,s & 3.2\,s & 227\,s \\
Keyframe localization & 1.4\,s & 1.4\,s & 176\,s \\
Inter-frame tracking (LK + PnP) & 23\,ms & 38\,ms & 79\,ms \\
\midrule
Effective throughput & 23.8\,FPS & 18.6\,FPS & 0.4\,FPS \\
\bottomrule
\end{tabular}
%\vspace{-0.5\baselineskip}
\end{table}

\subsection{Hyperparameter Sensitivity}
\label{sec:supp_hp}

\begin{figure}[ht]
	\centering
	\includegraphics[width=\linewidth]{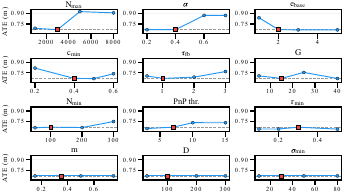}
	\caption{\textbf{Hyperparameter Sensitivity.} ATE vs.\ parameter value for all 12 hyperparameters (symbols in \cref{tab:hp_sensitivity}). Red square: default. Grey dashed: default ATE. Three parameters ($m$, $D$, $\sigma_{\min}$) show zero sensitivity, and the rest stay within 0.26\,m of the baseline.}
	\label{fig:hp_sensitivity}
	\vspace{-0.5\baselineskip}
\end{figure}

We perform a one-at-a-time sweep of 12 pipeline hyperparameters, evaluating 4 values per parameter across the 10 test sequences.
\cref{tab:hp_sensitivity} reports the maximal deviation from the default.
Three parameters ($m$, $D$, $\sigma_{\min}$) exhibit zero sensitivity, while the remaining nine produce only marginal variations ($\Delta$ATE$\leq 0.26$\,m). The most influential are the number of sampled matches $N$ and the point-drop ratio $\alpha$.
Median translation error and recall remain virtually unchanged ($\Delta$TE$\leq 0.03$\,m, $\Delta$R@2$\leq 0.9$\,percentage points), confirming that no parameter requires careful tuning.
\cref{fig:hp_sensitivity} visualizes the per-parameter trends.

\begin{table}[ht]
\centering
\caption{\textbf{Hyperparameter Sensitivity.} One parameter varied per row; others at default (bold). Symbol definitions in \cref{sec:supp_protocol}. $\Delta$ATE/$\Delta$TE: max.\ absolute deviation from default (m); $\Delta$R@2: in pp. Sorted by $\Delta$ATE.}
\label{tab:hp_sensitivity}
\setlength{\tabcolsep}{3pt}
\renewcommand{\arraystretch}{0.9}
\scriptsize
\begin{tabular}{@{}l c l rrr@{}}
\toprule
Parameter & Default & Swept Range & $\Delta$ATE & $\Delta$TE & $\Delta$R@2 \\
\midrule
$N$ & \textbf{3000} & [1000, 8000] & 0.26 & 0.03 & 0.9 \\
$\alpha$ & \textbf{0.4} & [0.2, 0.75] & 0.20 & 0.00 & 0.4 \\
$e_{\mathrm{base}}$ & \textbf{2} & [1.0, 5.0] & 0.17 & 0.01 & 0.5 \\
$c_{\min}$ & \textbf{0.4} & [0.2, 0.6] & 0.15 & 0.02 & 0.7 \\
$\tau_{\mathrm{fb}}$ & \textbf{1} & [0.5, 3.0] & 0.10 & 0.01 & 0.3 \\
$G$ & \textbf{15} & [5, 40] & 0.08 & 0.02 & 0.4 \\
$N_{\min}$ & \textbf{100} & [50, 300] & 0.08 & 0.01 & 0.3 \\
PnP reproj & \textbf{7} & [3.0, 15.0] & 0.06 & 0.02 & 0.6 \\
$r_{\min}$ & \textbf{0.3} & [0.1, 0.5] & 0.03 & 0.00 & 0.1 \\
$m$ & \textbf{0.35} & [0.15, 0.75] & 0.00 & 0.00 & 0.0 \\
$D$ & \textbf{100} & [30, 300] & 0.00 & 0.00 & 0.0 \\
$\sigma_{\min}$ & \textbf{30} & [10.0, 80.0] & 0.00 & 0.00 & 0.0 \\
\midrule
\multicolumn{3}{@{}l}{Default configuration (baseline)} & \multicolumn{3}{r}{ATE\,=\,0.67, TE\,=\,0.33, R@2\,=\,97.9\%} \\
\bottomrule
\end{tabular}
%\vspace{-0.5\baselineskip}
\end{table}

\section{MovingDrone Dataset Details}
\label{sec:supp_dataset}

This section provides additional details on the MovingDrone generation pipeline, realism augmentations, dataset splits, and statistics.

\subsection{Generation Pipeline}
\label{sec:supp_generation}

The pipeline converts a Google Earth Studio~\cite{googleearthstudio} camera trajectory into a complete multi-modal sequence in five stages.
(1)~Trajectory design: we design 194 camera paths across 21 regions in Google Earth Studio, following flight patterns from publicly available drone footage (orbital inspections, linear flyovers, oblique descents, and multi-turn paths). Google Earth Studio exports per-frame camera state including ECEF position, WGS84 coordinates, Euler angles, and vertical \ac{FOV}.
(2)~Speed normalization: the exported animations are cinematographic and often exceed 200~km/h, far beyond real drones. We resample the temporal parametrization via cubic splines to match the target speed distribution in \cref{tab:speed_dist} while preserving the spatial path. Position and rotation are interpolated jointly.
(3)~Rendering: we render photorealistic video from the Berlin VirtualCityMap~\cite{virtualcitymap_berlin} using Open3D offscreen EGL with $4\times$ MSAA at $1920\times 1080$. Depth maps use raycasting, guaranteeing pixel-exact depth-pose consistency without projection artifacts. All geometry is translated to a local frame to avoid floating-point cancellation from large UTM coordinates.
(4)~Geodata fetching: we automatically acquire reference data from the Berlin and Brandenburg open-data portals~\cite{geoportal_berlin,geoportal_brandenburg} based on each trajectory's UTM bounding box: multi-year \acp{DOP} at 0.2~m GSD (up to 14 years, 2011--2025, \cref{fig:supp_dop_all_years}), a \ac{DSM} at 0.2~m GSD, classified airborne \ac{LiDAR} (5--20~pts/m$^2$, ASPRS classes), and LoD1/LoD2 CityGML building models with semantic labels.
(5)~Coordinate alignment: we convert the exported coordinates to UTM Zone~33N. Each pose is stored as a camera center $(x, y, z)$ in UTM and a camera-to-world rotation quaternion $(q_w, q_x, q_y, q_z)$. A Z-offset calibration corrects the vertical datum difference between the rendering mesh and the \ac{DSM}.

\begin{table}[ht]
    \centering
    \caption{\textbf{Target Speed Distribution.} Proportions used for trajectory resampling.}
    \label{tab:speed_dist}
    \setlength{\tabcolsep}{4pt}
    \renewcommand{\arraystretch}{0.9}
    \scriptsize
    \begin{tabular}{llcc}
        \toprule
        Category & Reference & Speed (km/h) & Frac. \\
        \midrule
        Consumer slow & DJI Mini (filming) & 15--30 & 40\% \\
        Consumer cruise & DJI Mavic/Air & 30--50 & 30\% \\
        Prosumer max & DJI Mavic 3 & 50--72 & 20\% \\
        Racing / sport & FPV sport mode & 72--100 & 10\% \\
        \bottomrule
    \end{tabular}
%\vspace{-0.5\baselineskip}
\end{table}

\subsection{Realism Augmentations}
\label{sec:supp_augmentations}

Three physically motivated augmentations are applied to every sequence to narrow the sim-to-real gap.

\noindent\textbf{Trajectory Noise.}
Real \ac{UAV} platforms exhibit two noise regimes: slow GPS/wind-induced drift ($T_{\mathrm{slow}} \approx 3.5$~s) and fast motor micro-vibrations ($T_{\mathrm{fast}} \approx 0.1$~s). We generate two independent Gaussian-filtered signals per channel (weighted 30/70) and add them to the rendered trajectory. Roll banking proportional to yaw rate ($0.15^\circ$ per $^\circ$/frame) simulates coordinated turns. \Cref{tab:noise_params} lists per-channel amplitudes.

\noindent\textbf{Stochastic Jitter Events.}
We model two types of wind disturbances on top of the continuous noise above. Sudden gusts (20\% per-sequence probability) produce a 4-frame rotational spike of up to $0.8^\circ$. Sustained gusts (40\% probability) create 1--2~s episodes of amplified oscillation.

\noindent\textbf{Motion Blur.}
Directional motion blur is applied to frames where camera speed exceeds 15~m/s. A convolution kernel aligned with the projected 2D motion vector has length proportional to the effective speed $v_{\mathrm{eff}} = v + 0.1 \cdot \omega$ (translational + angular), clamped to $[3, 21]$ pixels. This affects 8--15\% of frames depending on the speed profile.

\begin{table}[ht]
    \centering
    \caption{\textbf{Trajectory Noise Parameters.}}
    \label{tab:noise_params}
    \setlength{\tabcolsep}{4pt}
    \renewcommand{\arraystretch}{0.9}
    \scriptsize
    \begin{tabular}{lcc}
        \toprule
        Channel & Amplitude ($\sigma$) & Bands \\
        \midrule
        Position XY & 0.30~m & drift + vibration \\
        Position Z & 0.12~m & drift + vibration \\
        Pitch, Yaw & $0.15^\circ$ & drift + vibration \\
        Roll & $0.15^\circ$ + $0.15 \cdot \dot\psi$ & drift + banking \\
        \bottomrule
    \end{tabular}
%\vspace{-0.5\baselineskip}
\end{table}

\subsection{Dataset Splits and Statistics}
\label{sec:supp_splits}

The four-way split (\cref{tab:dataset_stats}) disentangles two generalization axes: intra-region (novel viewpoints over familiar terrain) and inter-region (entirely unseen environments). Five geographically separated locations form the out-of-place test set, while the remaining 16 regions are divided into train, validation, and in-place test splits spanning diverse altitude and speed ranges. Although some \ac{DSM} tiles overlap between regions due to the tile grid, camera trajectories and viewpoints are entirely disjoint.

\begin{table}[ht]
    \centering
    \caption{\textbf{MovingDrone Split Statistics.} Mean~$\pm$~std of per-sequence values. Altitudes in \ac{AGL}. Range row: global extremes.}
    \label{tab:dataset_stats}
    \setlength{\tabcolsep}{2.5pt}
    \renewcommand{\arraystretch}{0.9}
    \scriptsize
    \begin{tabular}{lcccccccc}
        \toprule
        Split & Seq. & Regions & Frames & Speed (km/h) & Alt.~(m) & Traj.~(m) & Tilt ($^\circ$) & Dur.~(h) \\
        \midrule
        Train         & 174 & 16 & 297{,}518 & $46 \pm 23$ & $188 \pm 119$ & $545 \pm 295$ & $45 \pm 13$ & 2.8 \\
        Val           &  10 &  7 &  31{,}721 & $33 \pm 26$ & $302 \pm 283$ & $558 \pm 590$ & $38 \pm 21$ & 0.3 \\
        Test-InPlace  &   5 &  5 &   8{,}618 & $55 \pm 25$ & $216 \pm 84$  & $561 \pm 235$ & $46 \pm 8$  & 0.1 \\
        Test-OutPlace &   5 &  5 &   5{,}257 & $69 \pm 11$ & $207 \pm 104$ & $610 \pm 183$ & $46 \pm 6$  & 0.1 \\
        \midrule
        Total & 194 & 21 & 343{,}114 & $46 \pm 23$ & $195 \pm 134$ & $548 \pm 314$ & $44 \pm 14$ & 3.2 \\
        \cmidrule{4-8}
        Range & & & 300--12{,}473 & 0--116 & 26--1{,}145 & 139--2{,}289 & 0--83 & \\
        \bottomrule
    \end{tabular}
%\vspace{-0.5\baselineskip}
\end{table}

Individual sequences range from 10~s to 7~min at 30~fps (mean 1{,}769 frames). The three largest regions (Mitte, Potsdamer Platz, Airport) contribute 50\% of all frames. Per-sequence speeds closely follow the target distribution in \cref{tab:speed_dist}, and speed and altitude are weakly correlated, ensuring broad coverage of the flight-parameter space.

\subsection{Dataset Visualization}
\label{sec:supp_scene_diversity}

\begin{figure}[ht]
    \centering
    \includegraphics[width=\linewidth]{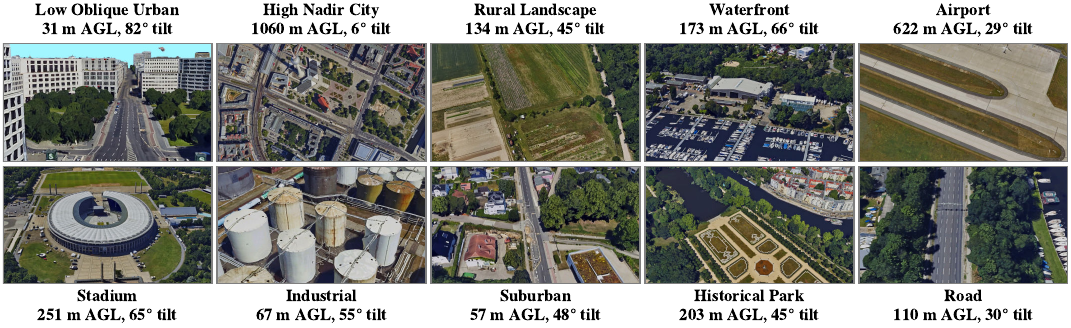}
    \caption{\textbf{Scene Diversity in MovingDrone.} 10 frames spanning 31--1{,}060\,m \ac{AGL}, nadir to highly oblique views, and different scene types.}
    \label{fig:dataset_diversity}
%\vspace{-0.5\baselineskip}
\end{figure}

\cref{fig:dataset_diversity} illustrates the diversity of MovingDrone with 10 representative frames spanning altitudes from 31 to over 1{,}000\,m \ac{AGL}, viewing angles from nadir to highly oblique, and nine scene types.

\begin{figure}[ht]
    \centering
    \includegraphics[width=\linewidth]{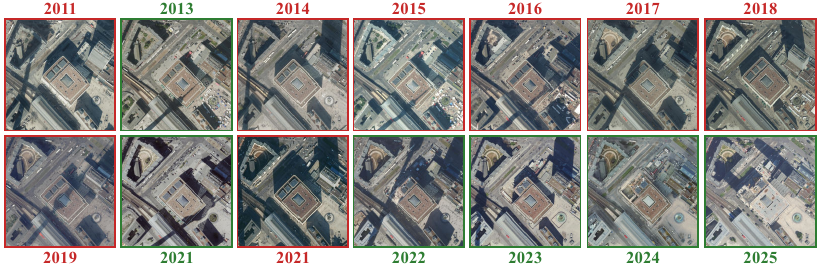}
    \caption{\textbf{All 14 DOP Years (2011--2025).} Same area around the Fernsehturm. Green borders: geometrically corrected TrueDOPs. Red borders: standard DOPs with visible building lean.}
    \label{fig:supp_dop_all_years}
\vspace{-0.5\baselineskip}
\end{figure}

\cref{fig:supp_dop_all_years} shows all 14 \ac{DOP} years (2011--2025) for the same area around the Fernsehturm. The building facades in standard \acp{DOP} highlights their perspective distortions compared to geometrically corrected TrueDOPs. The impact of \ac{DOP} vintage on localization is analyzed in \cref{sec:supp_dop_epoch}.

\begin{figure}[H]
	\centering
	\includegraphics[width=\linewidth]{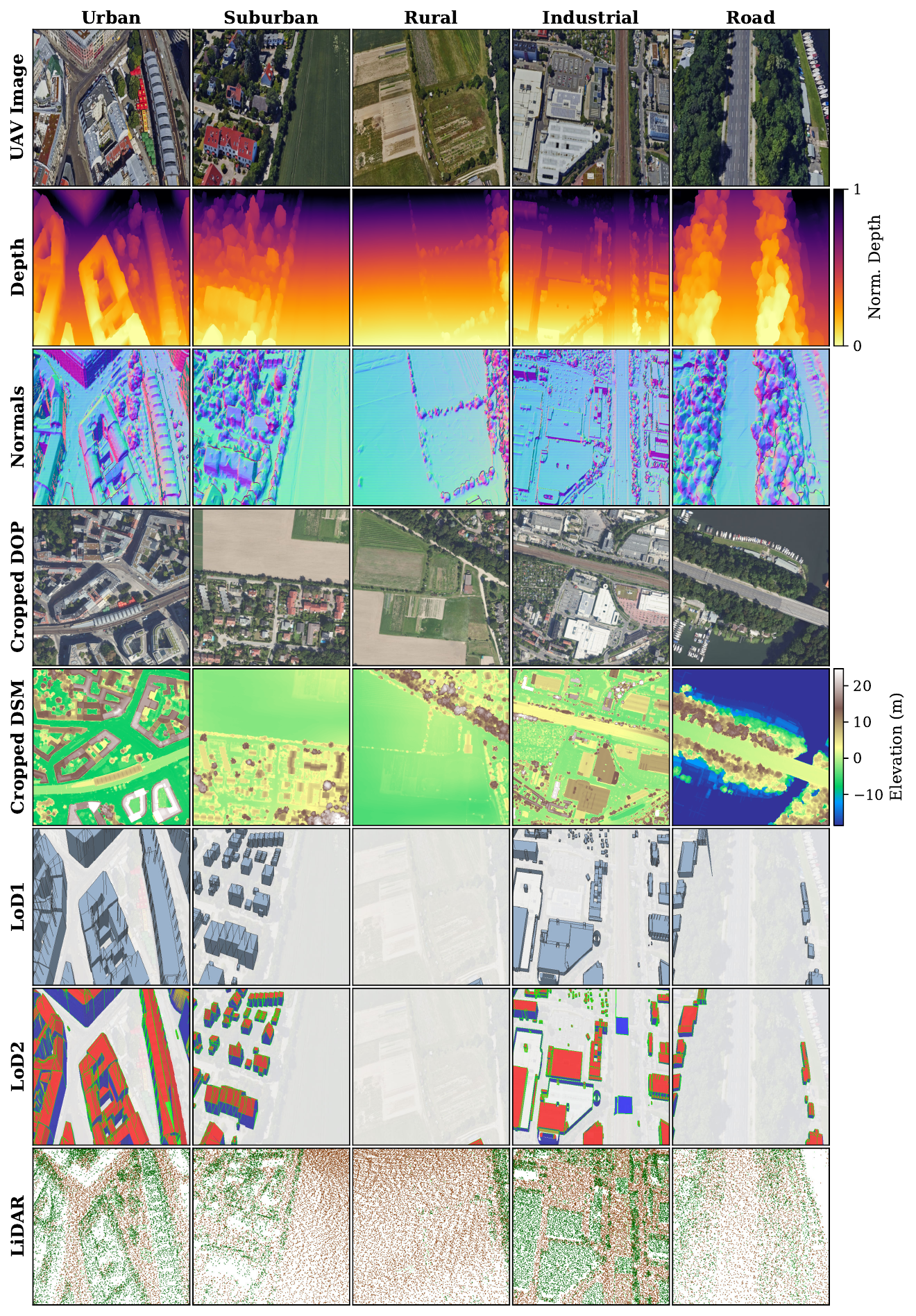}
	\caption{\textbf{Full-Modality Visualization.} Five sequences (columns) with all eight data modalities (rows), from \ac{UAV} query through depth, normals, \ac{DOP}, \ac{DSM}, LoD1/LoD2 meshes, and \ac{LiDAR}.}
	\label{fig:supp_sample}
	\vspace{-0.5\baselineskip}
\end{figure}

\cref{fig:supp_sample} presents all eight data modalities for five representative sequences. All spatial layers share the same UTM coordinate frame, enabling pixel-level cross-modal consistency.

\subsection{Licensing}
\label{sec:supp_license}

All geodata is published under open-data licenses by the Berlin Senate~\cite{geoportal_berlin} and Geobasis Brandenburg~\cite{geoportal_brandenburg}. The Berlin 3D city model VirtualCityMap~\cite{virtualcitymap_berlin} is available under the Data License Germany~2.0 (dl-de/by-2-0). Google Earth Studio~\cite{googleearthstudio} is used exclusively for camera trajectory design. No Google Earth imagery or renderings are included in the dataset, and all rendering is performed independently using the open-source VirtualCityMap mesh~\cite{virtualcitymap_berlin}. MovingDrone will be released with code for full reproducibility.

\section{Additional Experiments}
\label{sec:supp_experiments}

\subsection{Foundation Model Evaluation}
\label{sec:supp_foundation}

The comparison in main paper reports pose estimation results of foundation models under oracle Sim(3) alignment.
\cref{tab:supp_foundation_combined} additionally evaluates a first-frame and scale (ff+s) protocol that anchors the trajectory at the first ground-truth pose with a single global scale factor, simulating a scenario where an accurate initial 6-DoF pose and a global scale correction are available.
Since OrthoTrack produces metrically scaled absolute poses from geodata, its results are identical under both protocols.

Under oracle Sim(3), only DA3-Nested (ATE\,=\,2.7\,m) and DA3 (6.7\,m) reach single-digit ATE, yet both remain roughly $4{\times}$ worse than OrthoTrack (0.67\,m).
Switching to ff+s degrades all methods substantially, with the best ATE rising from 2.7 to 11.4\,m, because a single scale correction cannot compensate for rotational drift over long sequences.
DUSt3R-family methods produce rotation errors above $60^{\circ}$, confirming that pairwise global optimization is ill-conditioned for near-planar aerial geometry with small multi-view baselines relative to scene depth.

\begin{table}[ht]
	\centering\setlength{\tabcolsep}{3pt}
	\caption{\textbf{Foundation Model Evaluation under Two Alignment Protocols.} Each cell shows Sim(3)\,/\,ff+s values. \textbf{Bold}: best, \underline{underline}: second best.}
	\label{tab:supp_foundation_combined}
	\renewcommand{\arraystretch}{0.9}
	\scriptsize
	\begin{tabular}{l@{\hspace{4pt}}ccccc}
		\toprule
		Method & ATE$\downarrow$ & TE$\downarrow$ & RE$\downarrow$ & R@5$\uparrow$ & FPS$\uparrow$ \\
		\midrule
		DA3-Nested~\cite{lin2025depth}$\dagger$ & \underline{2.7}\,/\,\underline{11.4} & \underline{1.8}\,/\,\underline{10.8} & \underline{2.6}\,/\,\underline{1.1} & \underline{77.5}\,/\,\underline{63.5} & 11.3 \\
		DA3~\cite{lin2025depth}$\dagger$ & {6.7}\,/\,{28.1} & {4.0}\,/\,{27.4} & {3.4}\,/\,3.5 & {61.8}\,/\,13.9 & 21.5 \\
		Pi3~\cite{wang2025pi3}$\dagger$ & 24.0\,/\,72.9 & 19.3\,/\,66.5 & 39.5\,/\,31.1 & 30.0\,/\,18.9 & \textbf{25.2} \\
		Pi3X~\cite{wang2025pi3}$\dagger$ & 14.3\,/\,59.7 & 11.0\,/\,54.3 & 24.5\,/\,34.7 & 34.0\,/\,{23.7} & 14.2 \\
		VGGT-SLAM v2~\cite{wang2025vggt} & 11.9\,/\,39.6 & 9.3\,/\,37.8 & 15.9\,/\,{1.9} & 22.7\,/\,13.4 & 3.1 \\
		VGGT-SLAM v1~\cite{wang2025vggt} & 16.9\,/\,38.5 & 11.6\,/\,35.4 & 15.5\,/\,2.3 & 13.2\,/\,13.6 & 1.4 \\
		VGGT-Long~\cite{wang2025vggt}$\dagger$ & 28.6\,/\,67.1 & 27.0\,/\,53.8 & 17.5\,/\,2.3 & 20.7\,/\,12.7 & 1.0 \\
		VGGT~\cite{wang2025vggt}$\dagger$ & 31.4\,/\,93.0 & 27.4\,/\,64.9 & 26.5\,/\,12.8 & 12.9\,/\,10.9 & 15.8 \\
		MUSt3R~\cite{leroy2024grounding}$\dagger$ & 28.7\,/\,148.8 & 21.5\,/\,117.2 & 40.0\,/\,73.8 & 11.2\,/\,4.7 & 6.3 \\
		DUSt3R~\cite{wang2024dust3r}$\dagger$ & 53.4\,/\,139.8 & 43.6\,/\,112.2 & 81.3\,/\,89.9 & 4.5\,/\,5.1 & 0.2 \\
		MapAny.-K~\cite{keetha2025mapanything}$\dagger$ & 57.1\,/\,154.7 & 43.3\,/\,132.2 & 77.4\,/\,81.7 & 0.0\,/\,2.8 & 5.5 \\
		MapAny.~\cite{keetha2025mapanything}$\dagger$ & 55.7\,/\,145.4 & 43.8\,/\,114.5 & 94.5\,/\,105.6 & 0.0\,/\,2.7 & 5.5 \\
		PowR3R-BA~\cite{jang2025pow3r}$\dagger$ & 58.4\,/\,170.8 & 44.7\,/\,140.4 & 73.8\,/\,81.0 & 0.0\,/\,2.2 & 0.2 \\
		DUSt3R-M~\cite{leroy2024grounding}$\dagger$ & 55.4\,/\,118.2 & 47.3\,/\,86.9 & 62.2\,/\,69.6 & 0.0\,/\,3.1 & 0.2 \\
		MASt3R~\cite{leroy2024grounding}$\dagger$ & 130.6\,/\,256.2 & 114.3\,/\,234.2 & 126.5\,/\,116.5 & 0.0\,/\,1.1 & 1.5 \\
		\midrule
		OrthoTrack (Ours)$^\star$ & \textbf{0.67} & \textbf{0.33} & \textbf{0.06} & \textbf{90.9} & \underline{23.8} \\
		\bottomrule\end{tabular}\par
	\vspace{0.25\baselineskip}
	{\raggedright\scriptsize  ATE and TE in meters, RE in degrees, R@5 in \%. $\dagger$~windowed inference. $^\star$~Absolute poses (protocol-independent).\par}
	\vspace{-1\baselineskip}
\end{table}

\begin{figure}[H]
	\centering
	\includegraphics[width=\linewidth]{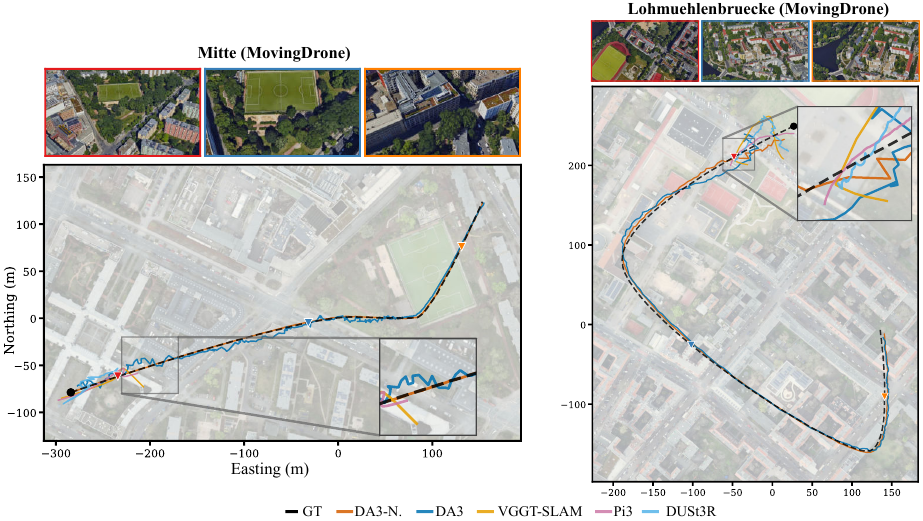}
	\caption{\textbf{Foundation Model Trajectories under Oracle Sim(3).} Five models on Mitte (left) and Lohmuehlenbruecke (right), overlaid on the \ac{DOP}. DA3-Nested closely tracks ground truth, while DUSt3R diverges due to rotation errors in near-planar aerial geometry.}
	\label{fig:supp_traj}
	\vspace{-0.5\baselineskip}
\end{figure}

\Cref{fig:supp_traj} shows predicted trajectories on two representative sequences under oracle Sim(3).
Top-performing methods remain close to the ground truth, while those with poor rotation estimation diverge significantly despite receiving the optimal global correction.

\subsection{Evaluation on UAVD4L}
\label{sec:supp_crossdataset}

UAVD4L~\cite{wu2023uavd4l} covers Changsha, China, with 19 sequences (${\sim}$3{,}800 frames) at 0.5\,fps, providing a geographically and operationally distinct test case.
Images (up to 5280$\times$3956\,px) are resized to a maximum dimension of 1{,}920\,px.
The benchmark provides only a textured mesh, from which we render the reference \ac{DOP} and \ac{DSM}, eliminating domain gap between geodata and the 3D scene model.

\begin{figure}[ht]
	\centering
	\includegraphics[width=\linewidth]{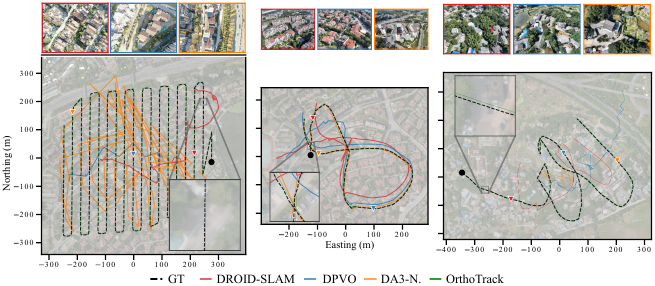}
	\caption{\textbf{UAVD4L Trajectory Visualization.} Despite oracle Sim(3), all baselines diverge, while OrthoTrack tracks the ground truth closely.}
	\label{fig:uavd4l_traj}
	%\vspace{-0.5\baselineskip}
\end{figure}

As shown in \cref{tab:uavd4l_comparison}, OrthoTrack localizes all 19 sequences with sub-degree rotation and near-perfect R@5.
Its slightly lower R@1 compared to DROID-SLAM reflects limited geometric accuracy of the mesh-derived \ac{DOP}/\ac{DSM}, where systematic offsets become the dominant error source at the 1\,m threshold.
Among foundation models, only DA3-Nested and Pi3 exceed DROID-SLAM's R@5, yet all remain far below OrthoTrack.
\cref{fig:uavd4l_traj} confirms that DROID-SLAM, DPVO, and DA3-Nested diverge substantially despite oracle Sim(3), while OrthoTrack closely follows the ground truth.
These results demonstrate that OrthoTrack also works in other scenes, cameras, and continents without adaptation, with accuracy scaling directly with geodata quality.

\begin{table}[H]
\centering
\caption{\textbf{Evaluation on UAVD4L~\cite{wu2023uavd4l}.} 19 sequences (6 inTraj + 13 outTraj). \textbf{Bold}: best; \underline{underline}: second best.}
\label{tab:uavd4l_comparison}
\setlength{\tabcolsep}{4pt}
\renewcommand{\arraystretch}{0.88}
\scriptsize
\begin{tabular}{l@{\hspace{4pt}}l@{\hspace{6pt}}rrrrrr}
\toprule
Method & Align. & ATE$\downarrow$ & TE$\downarrow$ & RE$\downarrow$ & R@1$\uparrow$ & R@2$\uparrow$ & R@5$\uparrow$ \\
\midrule
\multicolumn{8}{l}{\emph{VO / SLAM}} \\
Five-Point VO~\cite{nister2004efficient} & Sim(3) & 27.36 & 21.81 & 36.59 & 0.1 & 6.2 & 35.0 \\
ORB-SLAM3~\cite{orbslam3} & ff+s & 155.70 & 141.58 & 36.60 & 10.6 & 13.5 & 18.4 \\
ORB-SLAM3$^\ddagger$~\cite{orbslam3} & Sim(3) & 68.84 & 59.03 & 27.23 & 13.6 & 22.2 & 34.3 \\
DPVO~\cite{teed2024deep} & ff+s & 68.39 & 58.45 & 18.08 & 40.2 & 56.9 & 66.5 \\
DPVO~\cite{teed2024deep} & Sim(3) & 33.74 & 26.24 & 22.40 & \underline{57.0} & 66.1 & 68.4 \\
DROID-SLAM~\cite{teed2021droid} & ff+s & 68.70 & 56.35 & 21.39 & 40.3 & 62.0 & 68.3 \\
DROID-SLAM~\cite{teed2021droid} & Sim(3) & 32.47 & 24.70 & 19.09 & \textbf{62.7} & 67.9 & 68.5 \\
\midrule
\multicolumn{8}{l}{\emph{Foundation models}} \\
DA3-Nested~\cite{lin2025depth} & Sim(3) & 36.81 & 31.72 & 13.41 & 49.6 & 64.5 & 76.2 \\
DUSt3R$^{*}$~\cite{wang2024dust3r} & Sim(3) & 16.31 & 11.92 & 9.37 & 22.3 & 43.4 & 66.1 \\
Pi3$^{\S}$~\cite{wang2025pi3} & Sim(3) & \underline{5.67} & 4.43 & 6.32 & 41.8 & 61.2 & 73.5 \\
VGGT-SLAM~\cite{wang2025vggt} & Sim(3) & 47.94 & 39.76 & 11.13 & 24.8 & 37.9 & 50.1 \\
\midrule
\multicolumn{8}{l}{\emph{Geodata-based (no alignment)}} \\
OrthoLoC$^\P$~\cite{dhaouadi2025ortholoc} & --- & 1958.38 & \underline{1.09} & \underline{0.34} & 43.4 & \underline{84.8} & \underline{88.0} \\
OrthoTrack (Ours) & --- & \textbf{1.26} & \textbf{1.01} & \textbf{0.26} & 50.7 & \textbf{96.5} & \textbf{99.8} \\
\bottomrule
\end{tabular}\par
\vspace{0.25\baselineskip}
\parbox{\textwidth}{\raggedright\scriptsize Align.: Sim(3) = oracle 7-DoF, ff+s = first-frame anchor + scale, --- = absolute poses (no alignment). $^*$\,DUSt3R succeeded on 14/19 seq.\ (OOM on 5). $^\S$\,Pi3 on 12/19 (OOM on 7). $^\ddagger$\,ORB-SLAM3 Sim(3) from 16/19 (3 init failures). $^\P$\,OrthoLoC ATE dominated by sporadic catastrophic PnP failures; TE reflects typical-case accuracy.}
\vspace{-0.5\baselineskip}
\end{table}

\subsection{Challenging Scenarios}
\label{sec:supp_limitations}

\begin{figure}[H]
	\centering
	\includegraphics[width=\linewidth]{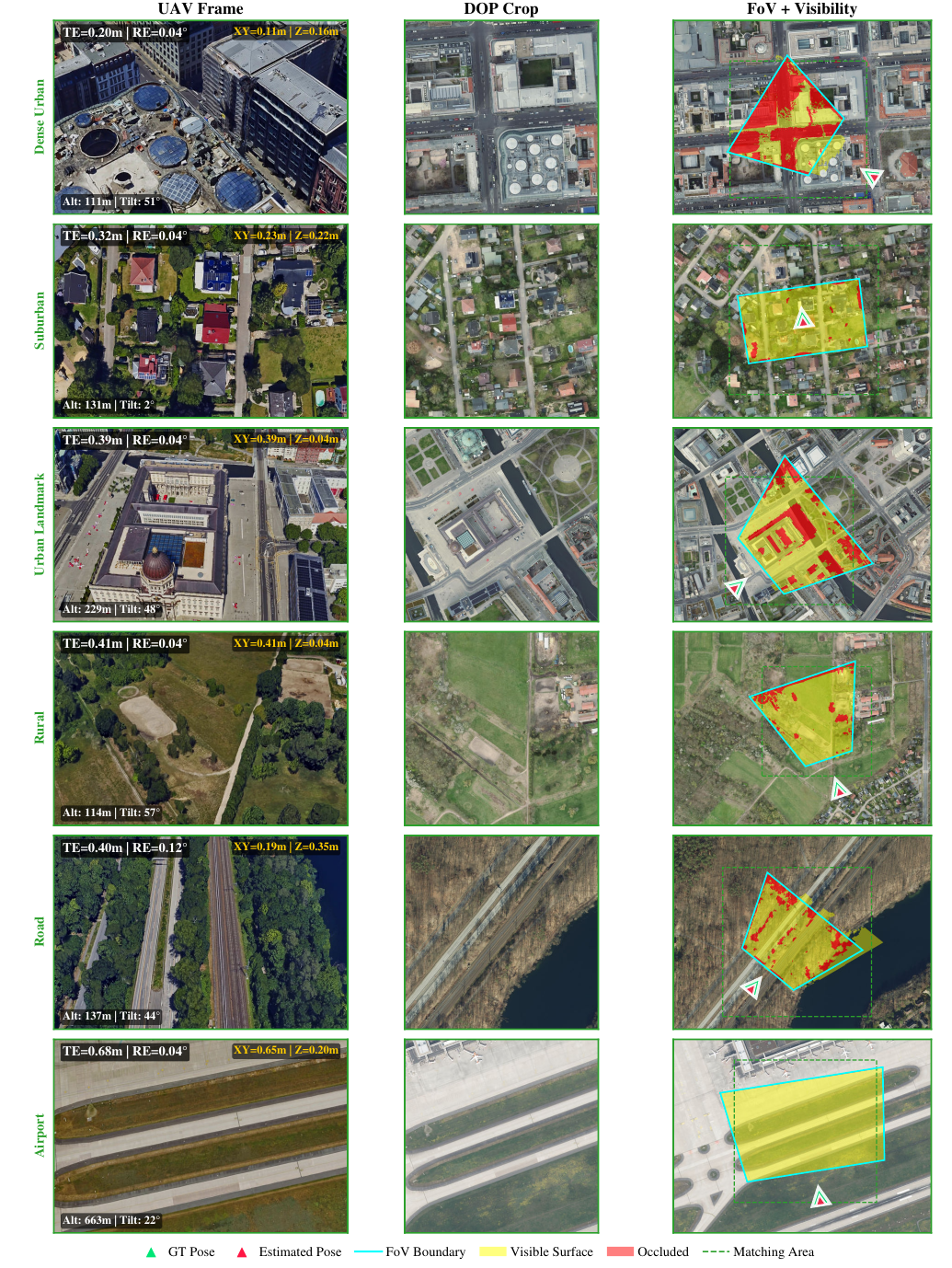}
	\caption{\textbf{Challenging Success Examples.} Same layout as \cref{fig:failure_modes}. Despite high tilt, varied altitude, and diverse environments, OrthoTrack achieves submeter accuracy in all cases.}
	\label{fig:success_examples}
	%\vspace{-0.5\baselineskip}
\end{figure}

\begin{figure}[H]
	\centering
	\includegraphics[width=\linewidth]{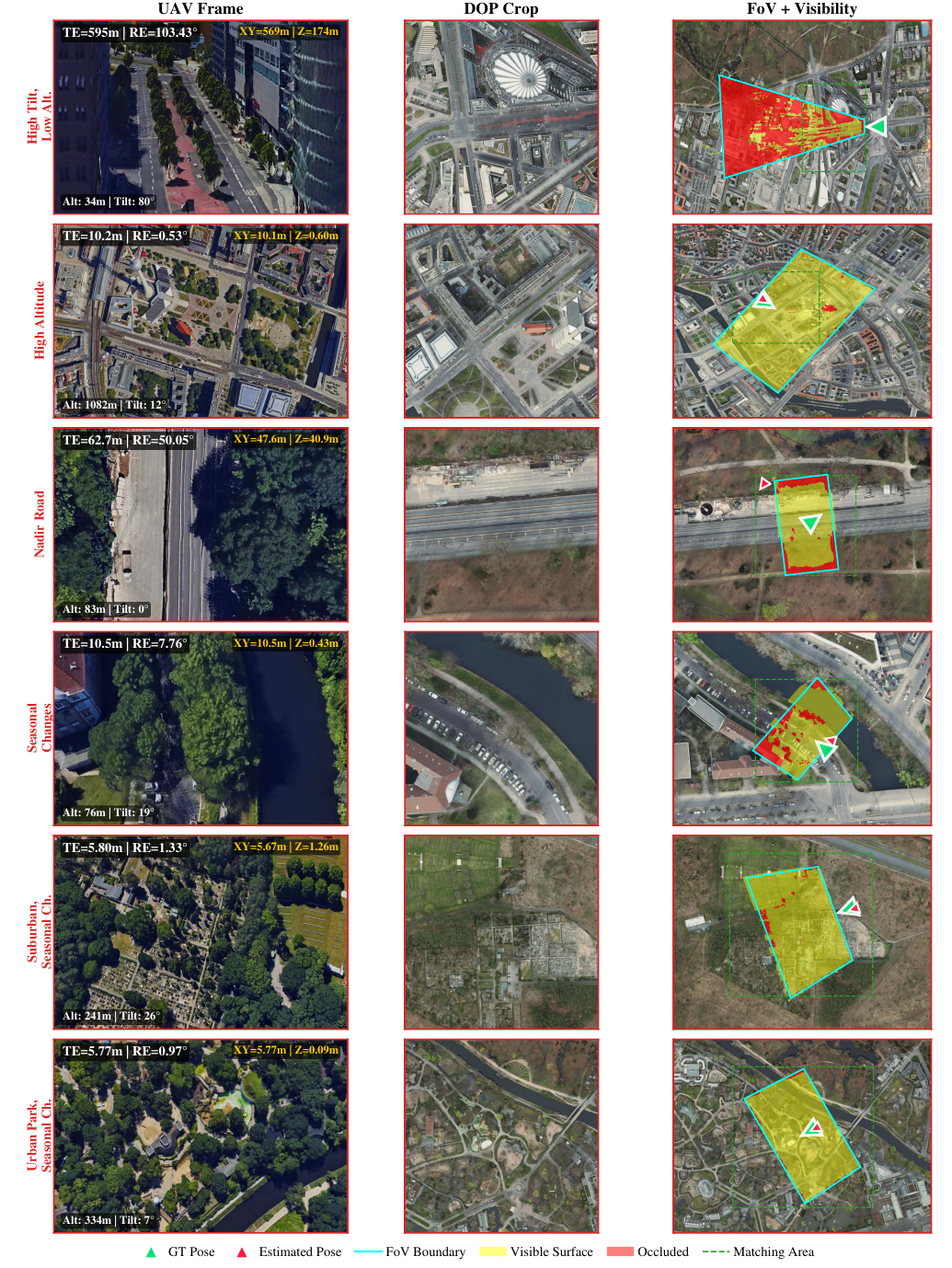}
	\caption{\textbf{Failure Mode Examples.} Each row: \ac{UAV} frame (left), \ac{DOP} crop (center), FoV overlay (right). Green/red triangles: GT/estimated positions. Per-frame TE, RE, altitude, and tilt annotated.}
	\label{fig:failure_modes}
	%\vspace{-0.5\baselineskip}
\end{figure}

\cref{fig:success_examples,fig:failure_modes} present challenging scenarios from MovingDrone to illustrate OrthoTrack's operating envelope and limitations.

\noindent\textbf{Success Cases.}
Dense urban scenes with distinctive architectural features achieve submeter accuracy across a wide range of conditions (\cref{fig:success_examples}), including highly tilted views, moderate and high altitudes, suburban near-nadir perspectives, road corridors, and airport environments. Performance is largely texture-dependent: regions with distinctive, stable structures support accurate matching regardless of altitude or tilt. OrthoTrack is robust to dynamic objects (PnP-RANSAC rejects outlier correspondences), temporal map change (submeter accuracy across 14 \ac{DOP} years, \cref{sec:supp_dop_epoch}), and transient failures (geodata-anchored matching prevents drift accumulation).

\noindent\textbf{Failure Cases.}
The most severe failures arise from extreme camera tilt ($>$70\textdegree{} from nadir), where the \ac{UAV} image shows building facades while the \ac{DOP} shows rooftops, creating a fundamental geometric mismatch (\cref{fig:failure_modes}). High-altitude nadir flights ($>$700\,m \ac{AGL}) suffer from GSD mismatch between the \ac{UAV} image and the 0.2\,m/px \ac{DOP}. Low-altitude road scenes fail due to repetitive lane markings and ambiguous correspondences from dynamic traffic. Suburban residential areas and parks with few distinctive features also produce occasional failures.

\noindent\textbf{Per-Frame Failure Rates.}
\cref{tab:failure_rates} quantifies OrthoTrack's per-frame failure rate, i.e., the fraction of frames where the pipeline does not produce a valid pose. Across all three benchmarks, the failure rate remains below 0.3\%, with the 10 MovingDrone test sequences achieving 0\% failures. The few failures concentrate in sequences with extreme tilt or low-texture areas.

\begin{table}[H]
\centering
\caption{\textbf{Per-Frame Failure Rates.} Frames where OrthoTrack fails to produce a valid pose. MovingDrone covers 35 sequences spanning the full altitude (26--1145\,m) and tilt (0--83\textdegree) range.}
\label{tab:failure_rates}
\setlength{\tabcolsep}{5pt}
\begin{tabular}{lccccc}
\toprule
Dataset & Seqs & Frames & Failed & Rate (\%) & Seqs w/ fail \\
\midrule
MovingDrone & 35 & 63,629 & 166 & 0.26 & 5/35 \\
MovingDrone (test) & 10 & 13,875 & 0 & 0.00 & 0/10 \\
UAVScenes & 20 & 120,591 & 8 & 0.01 & 2/20 \\
UAVD4L & 19 & 3,781 & 10 & 0.26 & 3/19 \\
\bottomrule
\end{tabular}
\end{table}

\begin{figure}[H]
	\centering
	\includegraphics[width=\linewidth]{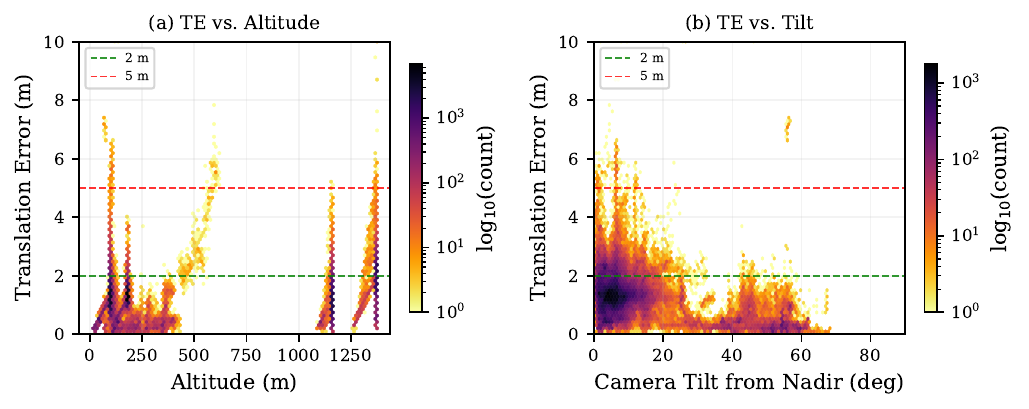}
	\caption{\textbf{Per-Frame Error Distribution.} Translation error vs. altitude (left, colored by tilt) and vs. camera tilt (right, colored by altitude) over 63k frames from 35 MovingDrone sequences. Green/red lines: 2\,m / 5\,m thresholds.}
	\label{fig:perframe_error_dist}
\end{figure}

\noindent\textbf{Per-Frame Error Distribution.}
\cref{fig:perframe_error_dist} shows the per-frame translation error as a function of altitude and camera tilt across 63{,}463 successfully localized frames from 35 MovingDrone sequences. Errors remain below 2\,m for the majority of frames up to 800\,m altitude and 60\textdegree{} tilt, establishing the operational bounds of the system. Accuracy degrades at high altitudes ($>$800\,m, median 2.73\,m) due to GSD mismatch and at extreme tilt ($>$60\textdegree, median 0.65\,m but with heavy-tailed outliers). The overall Pearson correlations are modest (TE-altitude: $r{=}{-}0.12$, TE-tilt: $r{=}0.20$), confirming that failures are driven by specific scene properties rather than altitude or tilt alone.

\subsection{Sensor Prior Noise Sensitivity}
\label{sec:supp_prior}

OrthoTrack does not require GPS or \acs{IMU} by default. The pipeline matches the first frame against the full downscaled \ac{DOP} to detect the approximate region, falling back to an exhaustive coarse-to-fine tile search if needed. All subsequent poses are determined solely by dense matching against the orthophoto.
When a coarse position prior is available (e.g., an initial GPS reading), it can optionally narrow the first-frame search region.
To quantify sensitivity, we sweep GPS horizontal noise $\sigma_h \in \{0, 1, 3, 5, 10, 20\}$\,m and \acs{IMU} yaw noise $\sigma_{\mathrm{yaw}} \in \{0^\circ, 2^\circ, \ldots, 30^\circ\}$ independently (\cref{tab:ablation_prior_noise}).

\begin{table}[ht]
\centering
\caption{\textbf{Sensor Prior Noise Sensitivity.} (a)~GPS noise sweep ($\sigma_{\mathrm{yaw}}$\,=\,4$^\circ$). (b)~IMU yaw sweep ($\sigma_h$\,=\,3\,m). No-prior default in \colorbox{gray!15}{grey}.}
\label{tab:ablation_prior_noise}
\setlength{\tabcolsep}{3pt}
\renewcommand{\arraystretch}{0.9}
\scriptsize
\begin{tabular}{l rrrrrr}
  \toprule
  Config & ATE$\downarrow$ & TE$\downarrow$ & RE$\downarrow$ & R@1$\uparrow$ & R@2$\uparrow$ & R@5$\uparrow$ \\
  \midrule
  \multicolumn{7}{l}{\textit{(a) GPS noise $\sigma_h$ (m)}} \\
  \quad$\sigma_h$\,=\,0     & 0.72 & 0.34 & 0.06 & 91.1 & 99.2 & 99.1 \\
  \quad$\sigma_h$\,=\,1     & 0.67 & 0.36 & 0.07 & 90.9 & 99.2 & 99.3 \\
  \quad$\sigma_h$\,=\,3     & 0.63 & 0.33 & 0.07 & 93.8 & 98.8 & 99.3 \\
  \quad$\sigma_h$\,=\,5     & 0.67 & 0.34 & 0.06 & 91.4 & 98.2 & 99.3 \\
  \quad$\sigma_h$\,=\,10    & 0.71 & 0.35 & 0.07 & 91.1 & 98.2 & 99.3 \\
  \quad$\sigma_h$\,=\,20    & 0.67 & 0.33 & 0.07 & 93.0 & 98.8 & 99.3 \\
  \midrule
  \multicolumn{7}{l}{\textit{(b) IMU yaw noise $\sigma_{\mathrm{yaw}}$ (deg)}} \\
  \quad$\sigma_{\mathrm{yaw}}$\,=\,0$^\circ$   & 0.66 & 0.34 & 0.06 & 92.4 & 98.8 & 99.3 \\
  \quad$\sigma_{\mathrm{yaw}}$\,=\,2$^\circ$   & 0.63 & 0.35 & 0.06 & 92.1 & 99.2 & 99.4 \\
  \quad$\sigma_{\mathrm{yaw}}$\,=\,4$^\circ$   & 0.63 & 0.33 & 0.07 & 93.8 & 98.8 & 99.3 \\
  \quad$\sigma_{\mathrm{yaw}}$\,=\,8$^\circ$   & 0.62 & 0.31 & 0.06 & 93.2 & 98.8 & 99.3 \\
  \quad$\sigma_{\mathrm{yaw}}$\,=\,15$^\circ$  & 0.64 & 0.35 & 0.07 & 91.9 & 99.3 & 99.4 \\
  \quad$\sigma_{\mathrm{yaw}}$\,=\,30$^\circ$  & 0.72 & 0.37 & 0.06 & 89.1 & 99.3 & 99.3 \\
  \midrule
  \rowcolor{gray!15}
  No prior           & 0.67 & 0.33 & 0.06 & 90.9 & 97.9 & 99.3 \\
  \bottomrule
\end{tabular}
%\vspace{-0.5\baselineskip}
\end{table}

The results show near-complete invariance: ATE varies by less than 0.1\,m across the full GPS noise range, and even 20\,m noise (far beyond typical consumer receivers) preserves submeter accuracy.
\acs{IMU} yaw noise follows a similarly flat trend.
Removing both priors entirely achieves essentially the same ATE as the prior-aided runs.
This robustness arises because the prior affects only the initial \ac{DOP} crop, and once the first match succeeds, subsequent frames rely on the tracked position.

\subsection{Geodata Sensitivity}
\label{sec:supp_geodata_sensitivity}
\begin{figure}[H]
	\centering
	\includegraphics[width=1\linewidth]{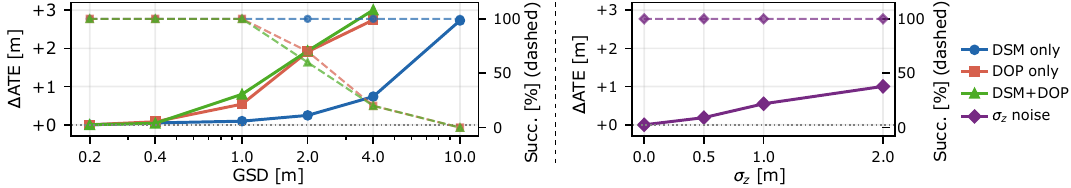}
    %\vspace{-1\baselineskip}
	\caption{\textbf{Geodata Sensitivity Sweep.} ATE as a function of \ac{DSM} and \ac{DOP} resolution (left) and \ac{DSM} noise $\sigma_z$ (right). \ac{DOP} resolution dominates the error, while \ac{DSM} noise is largely absorbed by PnP-RANSAC.}
    %\vspace{-0.5\baselineskip}
	\label{fig:supp_sweep}
\end{figure}

\noindent\textbf{Resolution}
To evaluate OrthoTrack's robustness to map quality, we sweep the \ac{DSM} and \ac{DOP} GSD, as well as \ac{DSM} noise $\sigma_z$ (\cref{fig:supp_sweep}).
The \ac{DOP} resolution dominates the error budget because it directly corrupts the 2D cross-view matches and thereby degrades $x,y,z$ coordinates. Conversely, a coarse \ac{DSM} perturbs only the $z$-coordinate during 3D lifting, and the PnP-RANSAC solver absorbs the resulting noisy correspondences, leading to only moderate localization errors even at elevated noise levels.

\noindent\textbf{Fallback for Missing Geodata.}
In real-world deployments, reference geodata may become unavailable or severely degraded. Such gaps can be bridged by falling back to relative \ac{SLAM} (e.g., DROID-SLAM), anchoring the trajectory to the last valid OrthoTrack pose and calibrating the scale using the most recent valid window. Experiments on the MovingDrone dataset demonstrate that for data gaps of 1s, 5s, 10s, and 15s, the mean ATE within the bridged segment is 0.52m, 0.63m, 0.86m, and 1.30m using online single-anchoring, and 0.54m, 0.36m, 0.46m, and 0.52m with offline two-end anchoring. This ensures trajectory continuity even during prolonged outages. Incorporating IMU measurements or 3D foundation models for local depth estimation could further enhance the robustness of these fallback mechanisms.

\subsection{Extended Matcher Ablation}
\label{sec:supp_matcher}

\cref{tab:supp_matcher_full} provides the complete matcher ablation referenced in main paper.
All six sparse matchers effectively fail (at most 10\% sequence success), confirming that local keypoint descriptors cannot bridge the aerial-to-orthophoto domain gap.
Among dense methods, L2M~\cite{liang2025learning}, GIM variants, MASt3R~\cite{leroy2024grounding}, and all RoMa v2 variants achieve 100\% sequence success. L2M attains the best per-frame TE and R@1, while RoMa v2 Precise achieves the best ATE.
Several recent dense matchers (DaD, EDM, RDD) reach at most 30\% sequence success despite strong standard benchmark results on MegaDepth and ScanNet, confirming that the aerial-to-orthophoto gap remains a distinct challenge.

\begin{table}[t]
\centering
\caption{\textbf{Extended Keyframe Matcher Ablation.} Only the keyframe matching backbone varies; LK flow tracker and PnP solver are shared. \textbf{Bold}: best; \underline{underline}: second best.}
\label{tab:supp_matcher_full}
\setlength{\tabcolsep}{4pt}
\renewcommand{\arraystretch}{0.9}
\scriptsize
\begin{tabular}{l@{\hspace{6pt}}rrrrrrr@{\hspace{6pt}}r}
\toprule
Matcher & ATE$\downarrow$ & TE$\downarrow$ & RE$\downarrow$ & R@1$\uparrow$ & R@2$\uparrow$ & R@5$\uparrow$ & Succ.$\uparrow$ & \#KF \\
\cmidrule(lr){1-9}
SP+SuperGlue$^*$~\cite{sarlin2020superglue} & 1.32 & 0.44 & 0.12 & 6.3 & 8.0 & 9.2 & 10\% & 2 \\
SP+LightGlue$^*$~\cite{lindenberger2023lightglue} & 0.91 & 0.41 & 0.09 & 7.7 & 9.3 & 10.0 & 10\% & 2 \\
DeDoDe~\cite{edstedt2024dedode} & -- & -- & -- & 0.0 & 0.0 & 0.0 & 0\% & 1 \\
XFeat+LightGlue$^*$~\cite{potje2024xfeat} & 3.48 & 0.67 & 0.17 & 6.2 & 7.2 & 8.8 & 8\% & 2 \\
RIPE~\cite{kunzel2025ripe} & -- & -- & -- & 0.0 & 0.0 & 0.0 & 0\% & 1 \\
LiftFeat~\cite{liu2025liftfeat} & -- & -- & -- & 0.0 & 0.0 & 0.0 & 0\% & 1 \\
\cmidrule(lr){1-9}
DKM~\cite{edstedt2023dkm} & 956.8 & 674.56 & 58.47 & 25.5 & 30.2 & 31.8 & 47\% & 15 \\
RoMa~\cite{edstedt2024roma} & 411.9 & 243.07 & 4.94 & 50.6 & 63.4 & 72.4 & \underline{95\%} & 14 \\
GIM(DKM)~\cite{wang2024efficient} & 2.56 & 0.52 & \underline{0.10} & 89.1 & \textbf{97.9} & \textbf{99.4} & \textbf{100\%} & 13 \\
GIM(RoMa)~\cite{wang2024efficient} & 4.63 & 1.16 & 0.18 & 46.9 & 84.6 & 96.2 & \textbf{100\%} & 13 \\
L2M~\cite{liang2025learning} & 6.23 & \textbf{0.29} & \textbf{0.06} & \textbf{94.9} & 96.4 & 98.3 & \textbf{100\%} & 15 \\
RoMa v2 (Precise) & \textbf{0.67} & \underline{0.33} & \textbf{0.06} & \underline{90.9} & \textbf{97.9} & \underline{99.3} & \textbf{100\%} & 15 \\
RoMa v2 (Base) & 0.90 & 0.49 & \underline{0.10} & 86.4 & 95.9 & 99.0 & \textbf{100\%} & 13 \\
RoMa v2 (Fast) & \underline{0.88} & 0.58 & 0.11 & 79.9 & \underline{96.9} & 99.1 & \textbf{100\%} & 13 \\
RoMa v2 (Turbo) & 4.66 & 1.60 & 0.29 & 47.4 & 73.0 & 91.3 & \textbf{100\%} & 12 \\
MASt3R~\cite{leroy2024grounding} & 5.74 & 2.10 & 0.46 & 20.4 & 56.3 & 85.6 & \textbf{100\%} & 11 \\
\cmidrule(lr){1-9}
RDD~\cite{chen2025rdd} & -- & -- & -- & 0.0 & 0.0 & 0.0 & 0\% & 1 \\
DaD(RoMa)$^*$~\cite{edstedt2025dad} & 0.96 & 0.68 & 0.10 & 22.8 & 27.7 & 30.0 & 30\% & 3 \\
EDM$^*$~\cite{li2025edm} & 4.08 & 3.55 & 0.35 & 9.8 & 10.0 & 10.0 & 10\% & 2 \\
UFM~\cite{zhang2025ufm} & 572.4 & 517.14 & 69.51 & 0.0 & 0.0 & 0.0 & 79\% & 149 \\
\bottomrule
\end{tabular}\par
\vspace{0.25\baselineskip}
\parbox{\textwidth}{\raggedright\scriptsize $^*$\,Metrics computed from the subset of sequences where the matcher produced valid poses. \#KF: median number of keyframe events per sequence. Succ.: percentage of sequences with at least one valid pose.}
\vspace{-1\baselineskip}
\end{table}

\subsection{DOP Year Analysis}
\label{sec:supp_dop_epoch}

\begin{table}[ht]
\centering
\caption{\textbf{DOP Year Ablation.} DSM fixed; all 14 years visualized in \cref{fig:supp_dop_all_years}.}
\label{tab:ablation_dop_year}
\setlength{\tabcolsep}{4pt}
\renewcommand{\arraystretch}{0.9}
\scriptsize
\begin{tabular}{l c rrrrrr}
  \toprule
  Year & True & ATE$\downarrow$ & TE$\downarrow$ & RE$\downarrow$ & R@1$\uparrow$ & R@2$\uparrow$ & R@5$\uparrow$ \\
  \midrule
  2025 & \checkmark & \textbf{0.67} & \textbf{0.33} & \textbf{0.06} & \textbf{90.9} & \textbf{97.9} & 99.3 \\
  2023 & \checkmark & 3.34 & \underline{0.44} & \underline{0.08} & \underline{88.5} & \underline{97.0} & \underline{99.4} \\
  2020 & \checkmark & \underline{0.93} & 0.65 & 0.15 & 78.4 & 94.5 & \textbf{99.9} \\
  2016 & & 70.55 & 26.28 & 0.78 & 14.6 & 27.9 & 54.3 \\
  2013 & \checkmark & 10.22 & 2.28 & 0.67 & 56.6 & 78.8 & 87.7 \\
  2011 & & 110.4 & 102.9 & 2.94 & 9.0 & 19.8 & 38.3 \\
  \bottomrule
\end{tabular}
%\vspace{-0.5\baselineskip}
\end{table}

\begin{figure}[H]
	\centering
	\includegraphics[width=\linewidth]{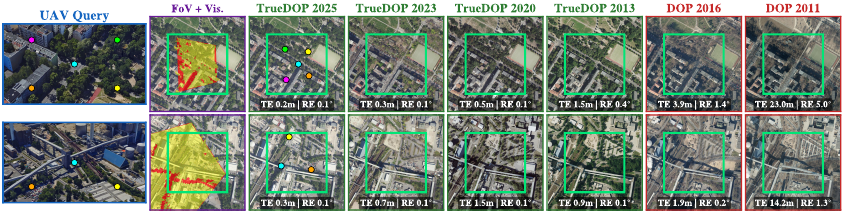}
	\caption{\textbf{DOP Temporal Appearance Gap.} Two sequences (rows) with \ac{UAV} query (blue border), camera \ac{FOV} overlay on the 2025 \ac{DOP} (yellow: visible, red: occluded), and corresponding \ac{DOP} crop per year (green bounding box).}
	\label{fig:supp_dop_uav_comparison}
	\vspace{-0.5\baselineskip}
\end{figure}

\cref{tab:ablation_dop_year} evaluates six \ac{DOP} vintages spanning 2011--2025 with a fixed \ac{DSM}.
Two factors dominate: geometric rectification and temporal appearance change.
Rectification has the largest effect: the 2013 TrueDOP achieves over $11{\times}$ lower TE than the newer 2016 standard \ac{DOP}, because perspective distortions near tall buildings corrupt 2D--3D correspondences (\cref{fig:supp_dop_all_years}).
Temporal appearance change affects all \acp{DOP}, including TrueDOPs: among the four TrueDOPs, newer vintages consistently achieve lower TE.
\cref{fig:supp_dop_uav_comparison} illustrates typical causes: seasonal vegetation (e.g., trees in full foliage obscuring road markings) and new construction alter scene appearance, degrading feature matching.
TrueDOPs with close temporal proximity should therefore be preferred.

\subsection{MovingDrone as a Monocular Depth Benchmark}
\label{sec:supp_depth}

Monocular depth estimation is a key enabling capability for \ac{UAV} applications such as obstacle avoidance, 3D mapping, and path planning, yet existing depth benchmarks focus almost exclusively on indoor and street-level scenes.

MovingDrone is uniquely suited for aerial depth evaluation: it provides pixel-level metric ground-truth depth rendered from the VirtualCityMap mesh via ray casting, covering a wide altitude range (26--1\,145\,m \ac{AGL}) and diverse urban, suburban, rural, and historic scenes.
We therefore use it to benchmark seven recent monocular depth estimators on the 10 test sequences.
We report two evaluation protocols in a single table (\cref{tab:depth_combined}): scale-aligned (median-ratio alignment per frame) and absolute (no alignment, testing whether the method's metric scale generalizes to aerial altitudes).

\begin{figure}[H]
	\centering
	\includegraphics[width=\linewidth]{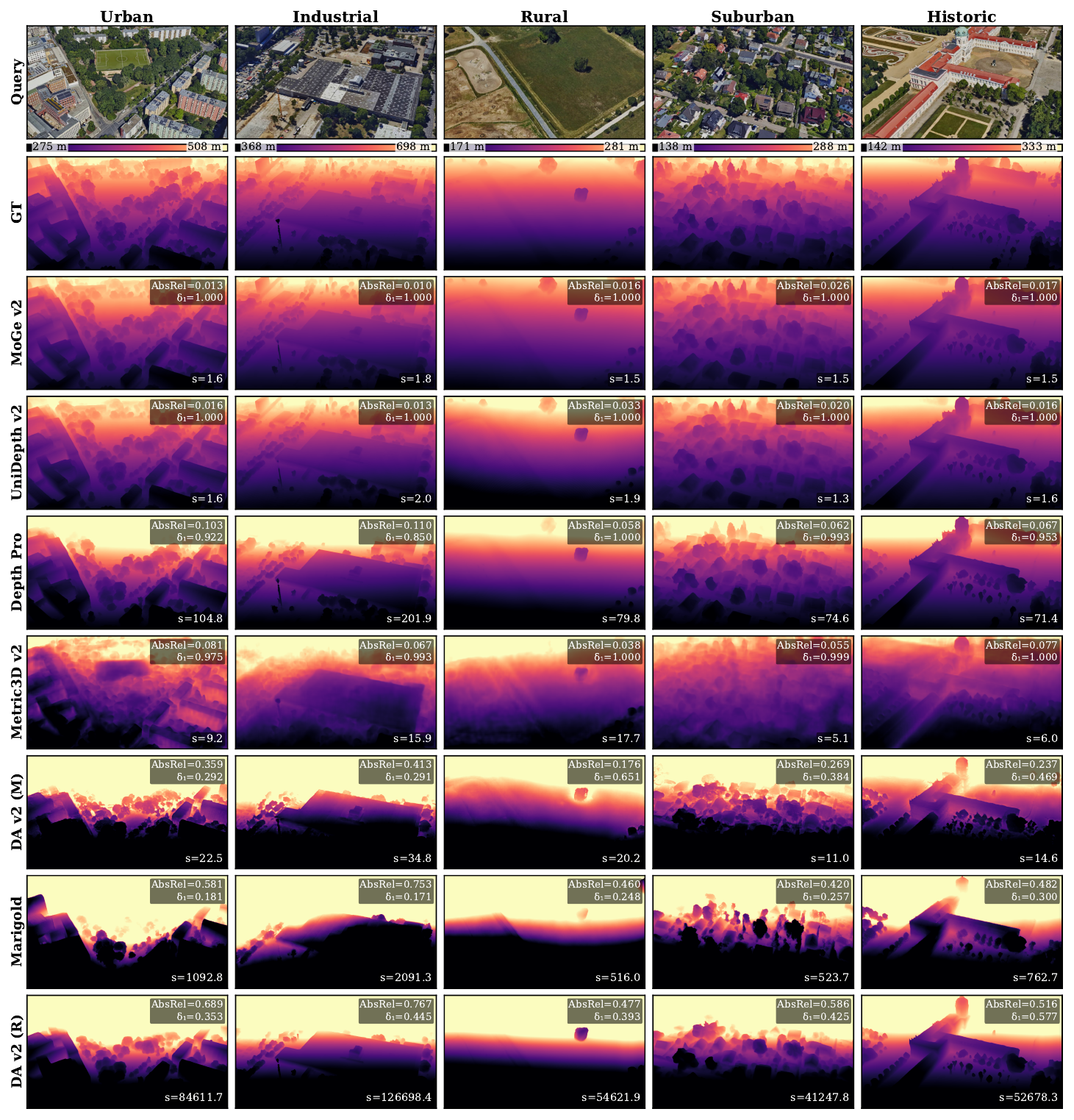}
	\caption{\textbf{Depth Estimation Qualitative Comparison.} Five scenes (columns) with GT depth and seven predictions (rows), all median-scale-aligned. MoGe~v2 and UniDepth~v2 match GT best, while DA~v2 and Marigold produce over-smoothed depth at aerial altitude.}
	\label{fig:supp_depth}
	%\vspace{-0.5\baselineskip}
\end{figure}

In the scale-aligned setting, MoGe v2~\cite{wang2025moge2} achieves near-perfect ordinal depth, followed by UniDepth v2~\cite{piccinelli2024unidepth}, as confirmed visually in \cref{fig:supp_depth}.
Diffusion-based models (Marigold~\cite{ke2025marigold}) and DepthAnything v2~\cite{yang2024da2} in relative mode perform poorly even with scale alignment, suggesting their ordinal ranking degrades for aerial viewpoints with extreme depth ranges.

\begin{table}[ht]
\centering
\caption{\textbf{Monocular Depth Estimation on MovingDrone.} Each cell: absolute\,/\,scale-aligned (median-ratio per frame). \textbf{Bold}: best; \underline{underline}: second best.}
\label{tab:depth_combined}
\renewcommand{\arraystretch}{0.95}
\scriptsize
\resizebox{\textwidth}{!}{%
\begin{tabular}{l@{\hspace{4pt}}c ccccccc r}
\toprule
Method & Type
  & AbsRel$\downarrow$ & SqRel$\downarrow$ & RMSE$\downarrow$ & logRMSE$\downarrow$ & $\delta_1\uparrow$ & $\delta_2\uparrow$ & $\delta_3\uparrow$ & FPS$\uparrow$ \\
\midrule
MoGe v2~\cite{wang2025moge2}
  & M
  & \textbf{0.30}\,/\,\textbf{0.014}
  & \underline{63.1}\,/\,\textbf{0.12}
  & \textbf{121.6}\,/\,\textbf{6.0}
  & \textbf{0.42}\,/\,\textbf{0.019}
  & \textbf{35.3}\,/\,\textbf{100.0}
  & \textbf{72.1}\,/\,\textbf{100.0}
  & \textbf{80.3}\,/\,\textbf{100.0}
  & \underline{4.6} \\
UniDepth v2~\cite{piccinelli2024unidepth}
  & M
  & \underline{0.39}\,/\,\underline{0.029}
  & \textbf{61.2}\,/\,\underline{0.53}
  & \underline{132.8}\,/\,\underline{11.8}
  & \underline{0.52}\,/\,\underline{0.036}
  & 12.1\,/\,\underline{99.8}
  & \underline{43.9}\,/\,\textbf{100.0}
  & \underline{71.3}\,/\,\textbf{100.0}
  & \textbf{5.6} \\
Depth Pro~\cite{bochkovskiy2025depthpro}
  & M
  & 0.99\,/\,0.082
  & 308.8\,/\,4.62
  & 315.5\,/\,38.0
  & 4.61\,/\,0.100
  & 0.0\,/\,94.9
  & 0.0\,/\,\underline{99.9}
  & 0.0\,/\,\textbf{100.0}
  & 0.3 \\
Metric3D v2~\cite{hu2024metric3d}
  & M
  & 0.90\,/\,0.093
  & 262.7\,/\,9.13
  & 290.5\,/\,38.8
  & 2.43\,/\,0.130
  & 0.0\,/\,93.6
  & 0.0\,/\,98.0
  & 0.0\,/\,\underline{98.8}
  & 1.3 \\
DA v2 (metric)~\cite{yang2024da2}
  & M
  & 0.94\,/\,0.311
  & 282.9\,/\,51.91
  & 301.1\,/\,131.4
  & 2.98\,/\,0.352
  & 0.0\,/\,37.3
  & 0.0\,/\,75.8
  & 0.0\,/\,97.3
  & 2.4 \\
\midrule
Marigold~\cite{ke2025marigold}
  & R
  & 1.00\,/\,0.458
  & 313.7\,/\,100.00
  & 318.0\,/\,179.6
  & 6.44\,/\,0.947
  & 0.0\,/\,27.6
  & 0.0\,/\,55.4
  & 0.0\,/\,74.2
  & 1.8 \\
DA v2 (rel.)~\cite{yang2024da2}
  & R
  & 0.62\,/\,0.542
  & 155.3\,/\,132.01
  & 218.7\,/\,199.8
  & 1.01\,/\,0.807
  & \underline{19.5}\,/\,23.3
  & 37.5\,/\,45.2
  & 52.7\,/\,64.7
  & 3.2 \\
\bottomrule
\end{tabular}}\par
\vspace{0.25\baselineskip}
{\raggedright\scriptsize M = metric estimator, R = relative-only. All values are per-sequence averages. $\delta_k$ thresholds follow the standard $1.25^k$ protocol.\par}
%\vspace{-0.5\baselineskip}
\end{table}

In absolute mode, all methods except MoGe v2 and UniDepth v2 produce near-zero $\delta_1$, indicating that current metric depth models are calibrated for indoor or street-level scales and fail to generalize to aerial altitudes without retraining.
Even the best absolute result (MoGe v2, AbsRel\,=\,0.30) is $20{\times}$ worse than its scale-aligned counterpart (0.014), highlighting the metric-scale gap as the dominant challenge for aerial depth estimation.

\section{Limitations and Future Work}
\label{sec:supp_future}

\noindent\textbf{Geodata Coverage.}
OrthoTrack leverages publicly available orthophotos and surface models, which are freely accessible for many countries worldwide. Our DOP year analysis (\cref{sec:supp_dop_epoch}) already shows robustness to outdated maps spanning 14 years. Extending the pipeline to additional map sources could further broaden coverage.

\noindent\textbf{Extreme Viewpoints.}
Highly tilted views ($>$70\textdegree) create a geometric mismatch between the UAV perspective and the nadir orthophoto, a challenge inherent to all orthophoto-based methods. OrthoTrack already handles typical aerial tilt ranges well (\cref{sec:supp_limitations}). Incorporating oblique imagery could further extend coverage to facade-level views.

\noindent\textbf{Future Directions.}
Since OrthoTrack is training-free and modular, it directly benefits from advances in dense matching without architectural changes. Faster or more viewpoint-invariant matchers would simultaneously improve accuracy and throughput, extending the current real-time capability.

\section{Ethical Considerations and Broader Impact}
\label{sec:supp_ethics}

\noindent\textbf{Privacy.}
OrthoTrack outputs only 6-DoF camera poses and performs no person detection, re-identification, or surveillance. MovingDrone is fully synthetic, rendered from the VirtualCityMap mesh~\cite{virtualcitymap_berlin} which models building geometry only and contains no people or identifiable content. All geodata is government-published under open-data licenses at 20\,cm/px, precluding individual identification.

\noindent\textbf{Dual Use and Safety.}
This project is developed with the explicit intent of advancing civilian UAV applications, such as search and rescue, disaster response, and infrastructure inspection. Our research aims to improve autonomy strictly for these positive societal use cases. The system relies on open-access geodata to complement commercial drones, and safe deployment requires task-specific validation.

\noindent\textbf{Bias and Environmental Footprint.}
MovingDrone reflects one European city (Berlin). Since OrthoTrack is training-free, generalization depends solely on the dense matcher. Experiments on UAVScenes and UAVD4L confirm transfer to non-European cities without adaptation. The pipeline runs in real time on a single GPU, keeping its footprint low.

\end{document}